\documentclass[10pt,journal,compsoc]{IEEEtran}

\ifCLASSOPTIONcompsoc
\else
\fi

\usepackage{microtype}
\usepackage{graphicx}
\usepackage{subfigure}
\usepackage{color}
\usepackage{booktabs}
\usepackage{enumitem}
\usepackage[table]{xcolor}
\definecolor{Gray}{gray}{0.9}
\definecolor{LightCyan}{rgb}{0.88,1,1}
\usepackage{amsmath,amssymb} % define this before the line numbering.
\usepackage{caption} % DO NOT CHANGE THIS AND DO NOT ADD ANY OPTIONS TO IT
\usepackage{hyperref}
\usepackage[american]{babel}
\usepackage{xspace}

\newcommand{\Last}{\method{LastShot}\xspace}

\usepackage{amssymb}
\usepackage{amsmath,amsfonts}
\usepackage{mathtools}
\usepackage{amsopn}
\usepackage{bm} % bold symbol
\usepackage{multirow}

\newcommand{\vct}[1]{\boldsymbol{#1}} % vector
\newcommand{\field}[1]{\mathbb{#1}}
\newcommand{\R}{\field{R}} % real domain
\newcommand{\ProbOpr}[1]{\mathbb{#1}}
\newcommand{\expect}[2]{%
\ifthenelse{\equal{#2}{}}{\ProbOpr{E}_{#1}}
{\ifthenelse{\equal{#1}{}}{\ProbOpr{E}\left[#2\right]}{\ProbOpr{E}_{#1}\left[#2\right]}}} % Expectation: syntax: E{1}{2} = E_1[2], E{}{2}=E[2], E{1}{} = E_1
\DeclareMathOperator{\argmin}{arg\,min}

\newcommand{\vtheta}{\vct{\theta}}
\newcommand{\vmu}{\vct{\mu}}

\newcommand{\vx}{{\vct{x}}}

\newcommand{\vphi}{\vct{\phi}}
\newcommand{\vpsi}{\vct{\psi}}

\newcommand{\eat}[1]{}
\newcommand{\method}[1]{\textsc{#1}}

\newcommand{\etal}{{\em et al.}}
\newcommand{\eg}{{\em e.g.}}
\newcommand{\ie}{{\em i.e.}}

\ifCLASSINFOpdf
\else
\fi

\begin{document}
	
\title{Few-Shot Learning with a Strong Teacher}

\author{Han-Jia Ye,
	Lu Ming,
	De-Chuan Zhan,
	Wei-Lun Chao
	
	\IEEEcompsocitemizethanks{
		\IEEEcompsocthanksitem H.-J. Ye, L. Ming, and D.-C. Zhan are with State Key Laboratory for Novel Software Technology,
		Nanjing University,  Nanjing, 210023, China; Wei-Lun Chao is with the Ohio State University, USA.
		\protect\\
		E-mail: \{yehj, mingl, zhandc\}@lamda.nju.edu.cn, chao.209@osu.edu}% <-this % stops an unwanted space
}

\markboth{Journal of \LaTeX\ Class Files,~Vol.~Xx, No.~X, Xxxx~20Xx}%
{Ye \MakeLowercase{\textit{et al.}}: Few-Shot Learning with a Strong Teacher}

\IEEEtitleabstractindextext{%
\begin{abstract}
Few-shot learning (FSL) aims to generate a classifier using limited labeled examples. Many existing works take the meta-learning approach, constructing a few-shot learner (a meta-model) that can learn from few-shot examples to generate a classifier. Typically, the few-shot learner is constructed or meta-trained by sampling multiple few-shot tasks in turn and optimizing the few-shot learner's performance in generating classifiers for those tasks. The performance is measured by how well the resulting classifiers classify the test (\ie, query) examples of those tasks. In this paper, we point out two potential weaknesses of this approach. First, the sampled query examples may not provide sufficient supervision for meta-training the few-shot learner. Second, the effectiveness of meta-learning diminishes sharply with the increasing number of shots (\ie, the number of training examples per class). To resolve these issues, we propose a novel meta-training objective for the few-shot learner, which is to encourage the few-shot learner to generate classifiers that perform like strong classifiers. Concretely, we associate each sampled few-shot task with a strong classifier, which is trained with ample labeled examples. The strong classifiers can be seen as the target classifiers that we hope the few-shot learner to generate given few-shot examples, and we use the strong classifiers to supervise the few-shot learner. We present an efficient way to construct the strong classifier, making our proposed objective an easily plug-and-play term to existing meta-learning based FSL methods. We validate our approach, \Last (\textbf{L}earning with \textbf{A} \textbf{S}trong \textbf{T}eacher for few-\textbf{SHOT} learning), in combinations with many representative meta-learning methods. On several benchmark datasets including {\emph{mini}ImageNet} and {\emph{tiered}ImageNet}, our approach leads to a notable improvement across a variety of tasks. More importantly, with our approach, meta-learning based FSL methods can consistently outperform non-meta-learning based methods at different numbers of shots, even in many-shot settings, greatly strengthening their applicability.
\end{abstract}
}

\maketitle

\IEEEpeerreviewmaketitle

%\IEEEraisesectionheading{\section{Introduction}}\label{sec:intro}

%!TEX root=main.tex
\section{Introduction}\label{sec:intro}
The intriguing ability that \emph{humans can learn new knowledge with few examples and then generalize it to novel environments} \cite{Lake2016BuildingMT,Lake2019HumanFL} has led to a community-wide enthusiasm in artificial intelligence towards few-shot learning (FSL)\footnote{The term \emph{``shot''} refers to the number of training examples per class.}, especially in the applications of 
visual recognition~\cite{sung2018learning,MotiianJID17Few}, machine translation~\cite{Gu2018Meta,Lake2019Compositional}, reinforcement learning~\cite{FinnAL17Model,FinnYZAL17One}, etc.  

Specifically for visual recognition, FSL is generally studied in a two-stage setting \cite{Wang2019SimpleShot,VinyalsBLKW16Matching,Chen2019Closer}. At first, the \emph{few-shot learner} --- a meta-model that aims to generate a classifier
given few-shot examples --- is provided with a large number of training examples collected from a set of \emph{base} classes. After learning from these data, the few-shot learner then receives a small number of training examples for a set of \emph{novel} classes that it needs to generate a classifier to recognize. The quality of the few-shot learner is measured by the generated classifier's accuracy on recognizing these \emph{novel} classes. By default, the two sets of classes are disjoint but related (\eg, both about animal species). Thus, to perform well in FSL, the few-shot learner must efficiently re-use what it learned (\eg, features) from the \emph{base} classes for the \emph{novel} classes.

\begin{figure}
	\centering
	\begin{minipage}[t]{0.49\linewidth}
		\centering
		\includegraphics[width=\textwidth]{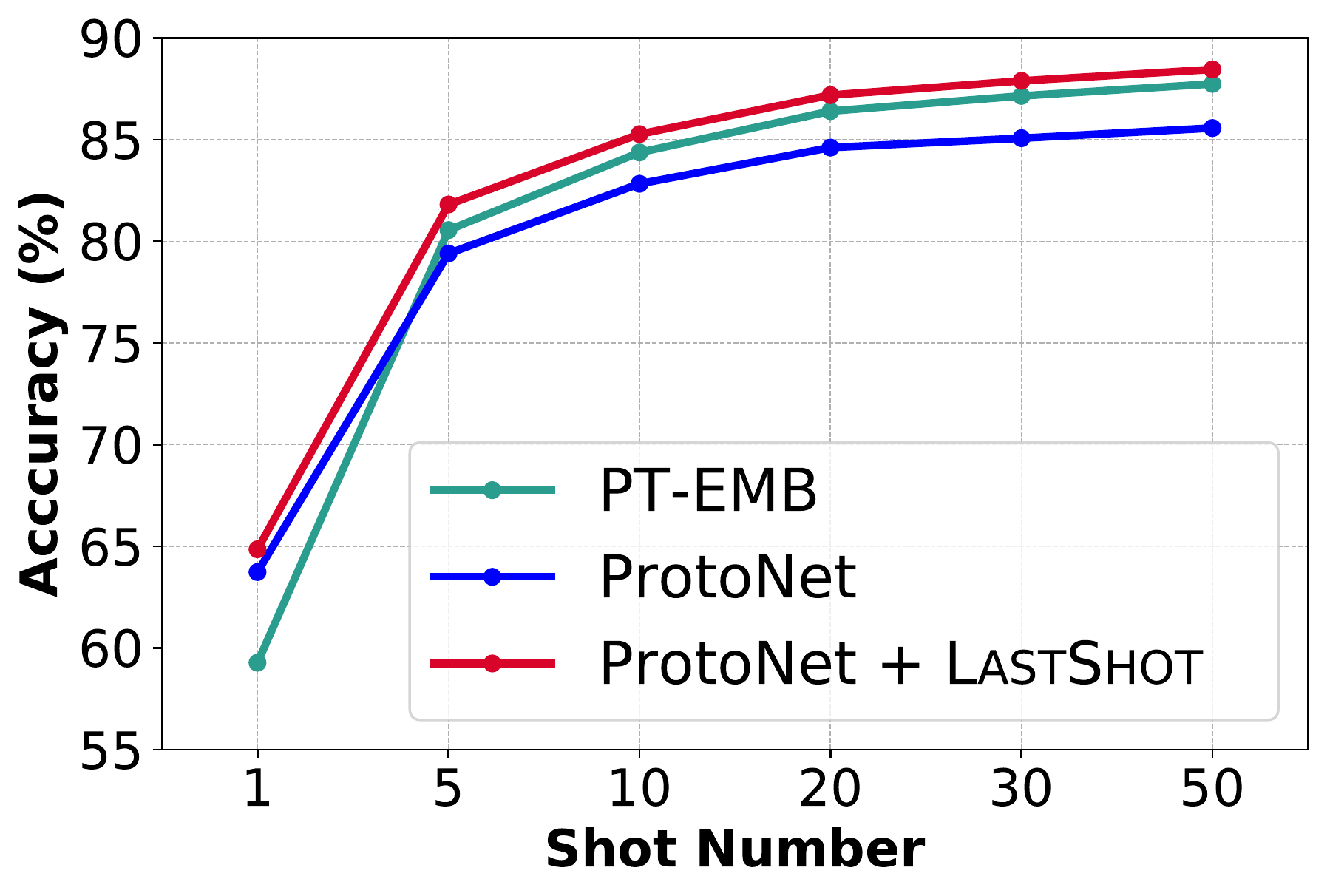}\\
	    \centering\mbox{(a) {\it mini}ImageNet}
	\end{minipage}
	\begin{minipage}[t]{0.49\linewidth}
		\centering
		\includegraphics[width=\textwidth]{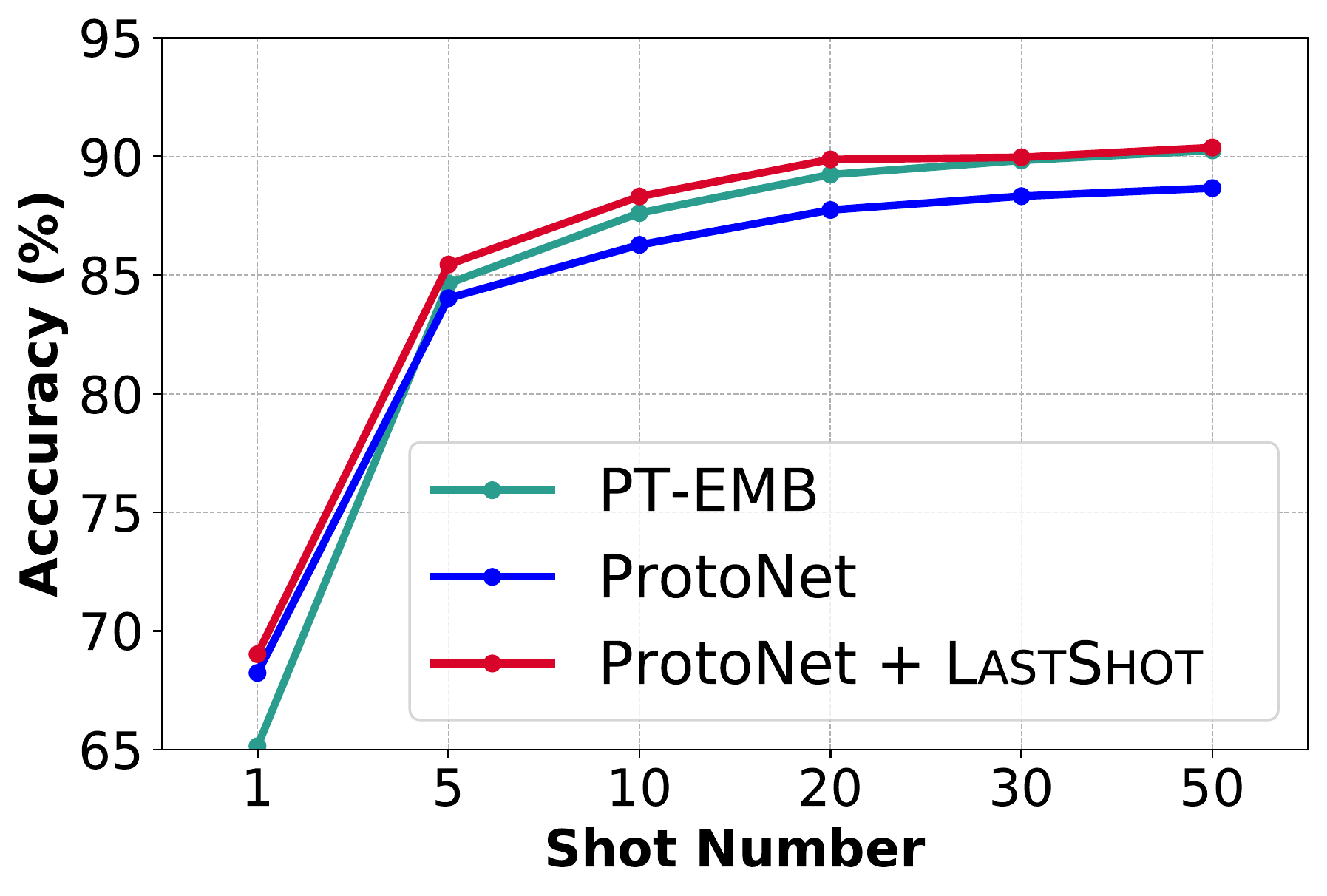}\\
	    \centering\mbox{(b) {\it tiered}ImageNet}
	\end{minipage}
	\caption{\textbf{The comparison of our \Last to existing methods.} We conduct 5-way (\ie, 5-class) classification experiments on {\it mini}ImageNet~\cite{VinyalsBLKW16Matching} and {\it tiered}ImageNet~\cite{ren2018learning} and report the accuracy under different numbers of shots.
	We consider two baselines: ProtoNet~\cite{SnellSZ17Prototypical} is a representative meta-learning based approach; PT-EMB is a nearest centroid classifier whose features are learned by directly training a multi-class classifier using all the base class data.
	ProtoNet outperforms PT-EMB on the 1-shot setting but falls behinds on the others.
    Our proposed {\Last} improves ProtoNet for all settings. See \autoref{sec:experiment} for details.}
	\label{fig:shot_change}
	\vskip -5pt
\end{figure}
 
Meta-learning, or learning to learn, is arguably the most popular approach to tackle FSL \cite{FinnAL17Model,AndrychowiczDCH16Learning,ravi2017optimization,SnellSZ17Prototypical,WangH16Learning}. The core idea is to \emph{simulate the few-shot scenario of learning with novel classes using the base class data.} That is, instead of using the {base} class data in a conventional way just like training a standard multi-class classifier \cite{he2016deep}, meta-learning approaches sample multiple \emph{few-shot tasks} from the {base} class data and use them to meta-train the few-shot learner to excel in the scenario it will encounter in receiving {novel} class data, {even though the {base} class data are indeed many-shot.} In other words, what the few-shot learner learns from the base class data is the ability to \emph{generate a classifier using few-shot data}.

To implement meta-learning for FSL, most existing works follow \cite{VinyalsBLKW16Matching} to formulate each simulated few-shot task by a support set and a query set whose data are sampled from the same subset of base classes. The support set is the few-shot training data that is used by the few-shot learner to generate the classifier; the query set is used to evaluate the resulting classifier's quality. In essence, the classification loss on the query set is the main supervision that is used to meta-train
the few-shot learner for how to learn from the support set.

In this paper, we argue that the classification loss on the query set is not an efficient form of supervision, especially when the query set size is limited. Concretely, in meta-training the few-shot learner, one usually constructs a mini-batch by sampling one or several few-shot tasks.
Thus, the limited query set size could result from the memory constraint or a large support set that consumes most of the space of a mini-batch\footnote{Conventionally, we sample one or several few-shot tasks (each includes a support set and a query set) whose size can fit into the mini-batch size (constrained by the memory). Thus, when the support set gets larger, either due to the increasing number of classes or shots, the query set size gets shrunk. In extreme cases, each class may just have one query example.}.
\autoref{fig:shot_change} gives an illustration, in which we experimented with different numbers of shots: the larger the shot number is for the support set, the smaller the query set size is.
We showed that, even in the standard 5-way 5-shot FSL setting, meta-learning methods (\ie, ProtoNet)~\cite{SnellSZ17Prototypical} can be outperformed by a simple nearest centroid classifier (\ie, PT-EMB)~\cite{Wang2019SimpleShot,mensink2013distance} whose features are trained with all the base class data in a conventional, non-meta-learning way. Moreover,
the gap between these two methods becomes more pronounced in a larger shot setting where the query set size shrinks\footnote{While one may overcome this by processing a few-shot task using multiple mini-batches, aggregating the gradients, and updating the model at once, this would increase the meta-training time and may involve additional hyper-parameter tuning.}.  
While one may argue that meta-learning should just be used in few-shot settings, the fact that it is effective only in a narrow shot range greatly limits its applicability.

The above issue motivates us to re-visit another implementation of meta-learning, which is to directly \emph{teach the few-shot learner to generate a strong classifier like the one trained using many-shot data} \cite{Chao2019Meta,Gui2018Few, WangH16Learning,WangRH17Learning}. In this way, the query set in a simulated few-shot task is replaced by a ``target'' classifier pre-trained using the (entire) base class data\footnote{Here, a target classifier means the classifier that we hope a few-shot learner can generate given few-shot examples. One choice according to \cite{Chao2019Meta,Gui2018Few, WangH16Learning,WangRH17Learning} is the classifier that is trained on ample labeled data.}. The loss for meta-training the few-shot learner is instantiated by a distance measure between the generated classifier and the target classifier in the parameter space. This implementation bypasses the limit of query set sizes. However, its applicability is limited by another factor: the lack of a suitable distance metric in the high-dimensional parameter space. Therefore, prior works pre-train a feature extractor in a non-meta-learning way, and constrain the few-shot learner to generate only the last fully-connected layer.

In this paper, we propose a novel meta-learning approach for FSL that incorporates the advantages of these two ways of implementations. On the high level, we follow the notion of constructing target classifiers to teach the few-shot learner, which bypasses the limit of query set sizes. 
However, instead of meta-training the few-shot learner to generate a classifier that is \emph{exactly like} the target classifier in the parameter space, 
we meta-train the few-shot learner to generate a classifier that \emph{predicts like} the target classifier on test (\ie, query) examples.
That is, in our approach, a simulated few-shot task contains both the target classifier and the query set besides the support set.
Given the support set, our approach encourages the few-shot learner to generate a classifier that would make predictions (\eg, predicted logits or class probabilities) on the query examples similar to the target classifier.
By using the difference in predictions rather than in the parameter space to supervise the few-shot learner, our approach enables the few-shot learner to learn to generate the entire classifier (\ie, the feature extractor plus the final fully-connected layer).
Moreover, measuring the difference in predictions disentangles the model architectures used by the target classifier, the few-shot learner, and the generated classifier, making our approach a plug-and-play meta-learning objective that is compatible with most existing meta-learning based FSL methods.
Even if the query examples are limited, the few-shot learner can receive extra and more informative supervision --- how a stronger target classifier makes predictions on them --- beyond the one-hot query example labels. 
 
We investigate multiple ways of constructing the target classifiers. One intuitive approach is to train a multi-class classifier using the entire base class data \cite{he2016deep}. However, since a simulated few-shot task usually contains only a subset of base classes (\eg, $5$ classes from the $64$ base classes \cite{VinyalsBLKW16Matching}), the prediction of this target classifier must be properly filtered and calibrated. We therefore propose to leverage this pre-trained classifier's feature extractor to (a) construct a target nearest centroid classifier \cite{Wang2019SimpleShot} or (b) dynamically train a target logistic regression, both using all data from the subset of base classes.
Besides, we further investigate multiple ways of querying the target classifiers' predictions \emph{during meta-training}. 
We found that imposing variations to the query data (\eg, resizing the images to be larger or adding noise) improves the performance, which is reminiscent of the recent studies in \cite{xie2020self}. During meta-testing on novel classes, we do not impose any variations to ensure a fair comparison to existing methods.
We name our approach \Last: \textbf{L}earning with \textbf{A} \textbf{S}trong \textbf{T}eacher for few-\textbf{SHOT} learning.

We validate \Last in combination with representative FSL methods \cite{SnellSZ17Prototypical,Ye2018Learning,triantafillou2019meta,Lee2019Meta} on multiple benchmarks, including \textbf{\emph{mini}ImageNet} \cite{VinyalsBLKW16Matching}, \textbf{\emph{tiered}ImageNet} \cite{Ren2018Meta}, \textbf{CIFAR-FS} \cite{bertinetto2018meta}, \textbf{FC-100} \cite{Oreshkin2018TADAM}, and \textbf{CUB} \cite{WahCUB_200_2011}. \Last consistently improves these methods and achieves the state-of-the-art accuracy. 
We further study \Last under varying numbers of shots in the support set. \Last shows highly robust performance, especially when the shots go larger and existing meta-learning based methods are outperformed by simple nearest centroid approaches \cite{Wang2019SimpleShot} (see \autoref{fig:shot_change}). The code in available at \url{https://github.com/Han-Jia/LastShot}.

The contributions of this work are as follow:
\begin{itemize}
    \item We propose a novel meta-training objective that encourages the few-shot learner to generate classifiers that perform like the corresponding target strong classifiers. Our proposed objective is an easily plug-and-play term to existing meta-learning based FSL methods.
    \item We present an efficient way to construct the target strong classifier.  While we meta-train the few-shot learner with few-shot tasks, we use the strong classifiers trained in a non-meta-learning way with ample examples to supervise the meta-training process.
    \item We conduct extensive experiments to validate our approach, which includes multiple benchmark datasets and multiple settings. Our \Last can consistently outperform conventional meta-learning and non-meta-learning based methods across a range of shot numbers.
\end{itemize}

\subsection{Comparisons to related work}
\label{ss_intro_comp}
On the surface, our approach is reminiscent of knowledge distillation \cite{hinton2015distilling}. However, there are several notable differences from other FSL methods that are based on knowledge distillation \cite{rajasegaran2020self,Zhang2020KnowledgeDF,wang2019progressive,Tian2929Rethinking}. For instance, Tian \etal~\cite{Tian2929Rethinking} studied how distillation with self-ensemble improves the pre-trained features for the downstream FSL task.
In contrast, we focus on how to fundamentally improve meta-learning based FSL methods by leveraging distillation in the meta-training process. That is, our approach is complementary to their efforts as we apply distillation at a different training stage (meta-training vs. feature pre-training)\footnote{Nowadays, nearly all meta-learning based FSL methods are initialized with pre-trained features learned in a conventional way with the base class data~\cite{Chao2019Meta,Ye2018Learning,Zhang2020Deep,Tian2929Rethinking}, and our approach is built upon them.}. Wang \etal~\cite{wang2019progressive} studied how to generate extra support set examples such that the generated classifier by the few-shot learner could perform similarly to the target classifier. In contrast, we do not introduce the extra data generation task, making our method more easily applicable. We also note that, our method is different from self-distillation~\cite{Bagherinezhad2018Label,Yang2019Snapshot,chen2020big,xu2020knowledge} --- our teacher is a classifier trained with ample data while our student is a few-shot learner. Please see \autoref{sec:related} for more details and discussions.
%!TEX root=main.tex
\section{Related Work}
\label{sec:related}

\subsection{Few-shot learning (FSL)}
FSL generally focuses on the $C$-way classification scenario in which the few-shot learner receives few labeled examples for $C$ new classes and needs to generate a classifier to recognize them ($C$ is usually set between $5$ to $30$). FSL can be studied from roughly three aspects from data, models, and algorithms~\cite{Wang2020Generalizing}. We describe non-meta-learning based FSL methods in this subsection, and describe meta-learning based methods in the next subsection.

From the aspect of data, FSL can be resolved if one has sufficient data for the $C$ novel classes.
\textbf{Data augmentation} approaches hallucinate additional examples of novel classes based on their few examples~\cite{antoniou2018data,hariharan2017low,Li2020Adversarial}. The hallucinated examples are used together with the original ones to train a better classifier for the novel classes. Since there are more training examples, it prevents the model from over-fitting.

From the aspect of models, FSL can be resolved if one can extract features that faithfully capture the semantic and discriminative information of each example to support nearest-neighbor classifiers.
\textbf{Feature or metric learning} approaches learn a feature extractor or metric that encourages examples from the same classes to be closer and examples from different classes faraway~\cite{mensink2013distance,koch2015siamese,schroff2015facenet}. The resulting features or metrics are used in a linear or nearest-neighbor classifier that is trained on the examples of the novel classes~\cite{Triantafillou2017Few,Scott2018Adapted,Wang2019SimpleShot}. Another direction to resolve FSL from the aspect of models is to construct \textbf{generative models}, which are learned to generate visual content conditioned on stochastic binary variables that indicate object identities. Recognizing novel classes involves adding new variables to the model and estimating the parameters associated with it based on few labeled data~\cite{fei2006one,lake2013one,edwards2017towards}.

From the aspect of algorithms, one could alter the optimization strategy to search for a better model given limited data. 
\textbf{Hypothesis transfer} reuses a well-trained model from a related dataset. Through fine-tuning it, we can leverage the knowledge learned from the related dataset to avoid over-fitting to few-shot data~\cite{Chen2019Closer,Dhillon2020Baseline}. \cite{Li2018Learning2,Ye2018Rectify,ye2021Heterogeneous} proposed special regularization strategies used during fine-tuning.

\subsection{Meta-learning for FSL} 
Here we mainly discuss how meta-learning helps FSL.
Meta-learning constructs simulated few-shot tasks, called episodes~\cite{VinyalsBLKW16Matching,ravi2017optimization}, from the base class data to mimic future few-shot tasks, and trains a few-shot learner to maximize the classification accuracy on the episodes. It is expected that the few-shot learner could be generalized to the few-shot tasks with novel classes. Various strategies have been investigated to implement the few-shot learner. For example, based on the meta-trained feature extractor, the few-shot learner classifies the query instances with (adapted) embeddings~\cite{VinyalsBLKW16Matching,SnellSZ17Prototypical,sung2018learning,Oreshkin2018TADAM,Li2019Few,Chen2021Learning,Pang2021Few}. Based on the meta-trained model initialization~\cite{FinnAL17Model,Lee2018Gradient,Nichol2018On,Yao2019Hierarchically,Rajeswaran2019Meta,Ye2021How}, the few-shot learner can adapt it to a good classifier for each few-shot task with a small number of gradient descent steps.
The optimization strategy for a few-shot learner to fit the few-shot examples can also be meta-trained and shared across episodes~\cite{bertinetto2016learning,rusu2019meta,Ren2019Incremental}, \eg, the learning rate~\cite{li2017meta,ravi2017optimization}. An image generator could also be meta-trained to augment the support set~\cite{Wang2018Low2}. 
We note that meta-learning can explicitly take characteristics of few-shot tasks into account. For instance, the number of novel classes $C$ and the number of labeled examples per class (\ie, shots) can be specified in the episodes.

Meta-learning based FSL has been applied in various fields, such as unsupervised learning~\cite{Hsu2018Unsupervised}, human motion prediction~\cite{Gui2018Few}, domain adaptation~\cite{Yue2021Prototypical}, image generation~\cite{Ojha2021Few}, and object detection~\cite{Juan2020Incremental,Zhu2021Semantic}. Some novel directions have also been explored recently.
For example, \cite{Liu2019Learning,liu2019learning2} leveraged the graph propagation to improve meta-learning; \cite{Gidaris2018Dynamic,Ye2021Learning} constructed a generalized FSL objective towards adaptive calibration strategies between base and novel classes; \cite{Yao2021Improving} investigated specialized data augmentations; \cite{Creager2021Environment} studied how to identify invariant features in meta-learning. Auxiliary self-supervised tasks are also incorporated to improve the generalization ability of the resulting few-shot learners or classifiers~\cite{Khodadadeh2019Unsupervised,Gidaris2019Boosting,Li2019Learning}.

Most of the meta-learning approaches rely on the classification loss on the query sets to drive the meta-training of the few-shot learner. \cite{WangH16Learning,WangRH17Learning,Qiao2018Few} were the first to propose the notion of constructing target classifiers to guide few-shot learners. Specifically, they measured the Euclidean distance between the target classifier and the classifier generated by the meta-learner as the loss. However, due to the large number of parameters, they only managed to meta-train the few-shot learner to generate the linear classifiers (\ie, the final fully-connected layers).

Some recent methods emphasize the importance of pre-trained features~\cite{koch2015siamese,rusu2019meta,Ye2018Learning,Chao2019Meta,Tian2929Rethinking,Zhang2020Deep}. In particular, the parameters of the backbone model are initialized by optimizing a classification objective over the base class data, which results in more discriminative features. Fine-tuning over the backbone leads to strong FSL baselines~\cite{Chen2019Closer,Dhillon2020Baseline}.
The influence of the number of shots (\eg, between meta-training and meta-testing) is discussed in~\cite{Ravichandran2019Few,Cao2020Theoretical,Oreshkin2018TADAM}.

We empirically find that the effectiveness of meta-learning diminishes sharply with increasing shots. Specifically, compared to the simple classifiers built upon the pre-trained embeddings, the episodic meta-training could not get further improvement. 
We propose a plug-and-play module \Last for meta-learning based FSL to encourage the few-shot learner to generate classifiers which perform like strong classifiers. This helps the resulting few-shot learner outperform other methods across different shot numbers. 

\subsection{Knowledge distillation}
As mentioned in \autoref{ss_intro_comp}, our approach is reminiscent of knowledge distillation \cite{hinton2015distilling} as we leverage target classifiers trained on ample examples to supervise the meta-training of the few-shot learner.
In this subsection, we therefore review knowledge distillation and contrast our usage of it to other existing methods for few-shot learning \cite{rajasegaran2020self,Zhang2020KnowledgeDF,wang2019progressive,Tian2929Rethinking}.

\textbf{Knowledge distillation (KD)}  \cite{hinton2015distilling} is a training strategy that leverages the dark knowledge from one (teacher) model to facilitate the learning progress of another (student) model. One popular way to extract the dark knowledge is by the instance-wise prediction, \eg, the posterior probability on a data instance~\cite{hinton2015distilling,SauB16,Mirzadeh2019Improved}, which provides the relative comparison among classes, in contrast to the conventional one-hot label. By matching the predictions of a ``student''  model to those of a ``teacher'' model, the student model obtains richer learning signals and becomes more generalizable. 
The dark knowledge for KD can be in the form of soft labels~\cite{hinton2015distilling}, hidden layer activation~\cite{RomeroBKCGB14,Koratana2019LIT}, and comparison relationships~\cite{Ye2020Distilling}.
Knowledge distillation has also been exploited for model interpretability~\cite{ZhouJ04} and incremental learning~\cite{Li2018Learning,Tao2020Topology}, etc.

Besides~\cite{Tian2929Rethinking,wang2019progressive} mentioned in \autoref{ss_intro_comp}, KD was also employed in two recent works of FSL~\cite{rajasegaran2020self,Zhang2020KnowledgeDF}, but again based on very different motivations and manners from our \Last. \cite{Zhang2020KnowledgeDF} applied KD mainly to construct a portable few-shot learner; \cite{rajasegaran2020self} applied self-supervised learning \cite{jing2020self,liu2020self} to train robust features and use them to guide the few-shot learner. Both of them only studied a single FSL method, \eg, \cite{FinnAL17Model}. In contrast, our approach is more general, compatible with many meta-learning based FSL methods and requiring no self-supervised learning steps. 
Wang~\etal~\cite{wang2019progressive} investigated constructing a data generator to create extra support set examples; the generator is learned via KD, from a many-shot target classifier.
In \Last, we take advantage of a plug-and-play loss term to distill knowledge from the target classifier more easily.

\textbf{Self-distillation} is a special case of KD which distills knowledge from the model itself, \ie, the teacher can be the previous generation of the model along the training progress~\cite{Bagherinezhad2018Label,FurlanelloLTIA18,Yang2019Snapshot}. 
Tian \etal~\cite{Tian2929Rethinking} applied self-distillation in the model pre-training stage to obtain more discriminative features. 
Our \Last works in an orthogonal direction to directly improve the meta-learning based FSL methods. 
Different from self-distillation, the target classifier in \Last is not the previous few-shot learner or feature extractor, but a stronger classifier trained with ample base class data.
%!TEX root=main.tex
\section{Our Approach: \Last}
\label{sec:approach}

In this section, we first introduce the notation of few-shot learning (FSL) for visual recognition and highlight the overall framework of meta-learning for FSL. We then introduce our approach \Last.

\subsection{Notation}
\label{ss_notation}
In FSL, a few-shot learner is first provided with a training set $\mathcal{D}_\text{base} = \{(\vx_1, y_1), \dots, (\vx_N, y_N)\}$ that contains $N$ labeled images from $B$ base classes; \ie, $y_n\in\{1,\cdots, B\}$. Here, each class is supposed to have ample training examples.
The few-shot learner is then provided with a support set $\mathcal{D}_\text{support}'$ of labeled images from $C$ novel classes $\{B+1,\cdots, B+C\}$, in which each novel class has $K$ examples. 
The goal of the few-shot learner is to generate a classifier that can accurately classify among the $C$ novel classes. This FSL setting is referred to as the $C$-way $K$-shot setting. 

\subsection{Meta-learning for few-shot learning}
\label{ss_basic_meta}
The core concept of meta-learning is to use the base class data
$\mathcal{D}_\text{base}$
to simulate the FSL scenario for meta-training the few-shot learner, before it receives the few-shot data from the novel classes. 
Namely, meta-learning samples multiple FSL tasks $\mathcal{T}$ from $\mathcal{D}_\text{base}$, each contains a $C$-way $K$-shot support set $\mathcal{D}_\text{support}$ whose class labels are from $\{1,\cdots, B\}$. Meta-learning then inputs each support set to the few-shot learner $\mathcal{A}$, whose goal is to output/generate a classifier that can classify the $C$ input classes well. To evaluate the quality of $\mathcal{A}$, meta-learning approaches usually sample a corresponding query set $\mathcal{D}_\text{query}$ for each $\mathcal{D}_\text{support}$, whose examples are from the same $C$ classes. The quality of $\mathcal{A}$ is measured by the outputted classifier's accuracy on $\mathcal{D}_\text{query}$.

Putting all these notations together, we can now formulate the meta-learning problem as follows
\begin{align}
    \min_\mathcal{A} \sum_{\mathcal{T}\sim\mathcal{D}_\text{base}} \;\sum_{(\vx,y)\in\mathcal{D}_\text{query}} \ell(\mathcal{A}(\mathcal{D}_\text{support})(\vx), y), \label{eq_meta}
\end{align}
where each $\mathcal{T}$ is a tuple $(\mathcal{D}_\text{support},\mathcal{D}_\text{query})$; $\mathcal{A}(\mathcal{D}_\text{support})$ is the $C$-way classifier outputted by $\mathcal{A}$, after taking the support set $\mathcal{D}_\text{support}$ as input. $\ell$ is a loss function on the prediction; \eg, the cross-entropy loss. As shown in \autoref{eq_meta}, the query set $\mathcal{D}_\text{query}$ is the major supervision that drives the meta-training of the few-shot learner $\mathcal{A}$.

After $\mathcal{A}$ is properly meta-trained, we then apply it to the \emph{true} $\mathcal{D}_\text{support}'$ of the novel classes to construct the classifier.

\subsubsection{Exemplar few-shot learners} \label{ss_meta-examplar} We review several representative few-shot learners. In model-agnostic meta-learning (MAML) \cite{FinnAL17Model}, the few-shot learner $\mathcal{A}$ is modeled by a neural network classifier $f_{\vtheta}$ with weights $\vtheta$. Denote by $\mathcal{L}(\vtheta)$ the differentiable loss of $f_{\vtheta}$ computed on the support set $\mathcal{D}_\text{support}$,  $\mathcal{A}(\mathcal{D}_\text{support})$ is formulated, in its simplest form, as an updated neural network $f_{\hat{\vtheta}}$ with weights $\hat{\vtheta}$:
\begin{align}
\hat{\vtheta}  = \vtheta - \alpha \times \nabla_{\vtheta} \mathcal{L}(\vtheta),\label{e_MAML_inner}
\end{align}
where $\alpha$ is a step size.
That is, MAML performs a single (or few) gradient descent update(s) according to $\mathcal{L}(\vtheta)$ to obtain the outputted classifier. The meta-learning problem in \autoref{eq_meta} thus aims to find a good initialization $\vtheta$ such that the neural network $f_{\hat{\vtheta}}$ can perform well on $\mathcal{D}_\text{query}$ after only a single (or few) update(s). Many meta-learning approaches are built on MAML \cite{Lee2018Gradient,finn2018meta,li2017meta,rusu2019meta,boney2018semi,grant2018recasting,triantafillou2019meta,Lee2019Meta}.
  
Another popular way to model $\mathcal{A}$ is via a feature extractor $f_{\vtheta}$. For example, in Prototypical Network (ProtoNet)~\cite{SnellSZ17Prototypical}, $\mathcal{A}(\mathcal{D}_\text{support})(\vx)$ is formulated as 
\begin{align}
\hat{y} = \argmin_c \left\|f_{\vtheta}(\vx) - \sum_{\vx'\in\mathcal{D}_\text{support, c}} \frac{f_{\vtheta}(\vx')}{|\mathcal{D}_\text{support, c}|}\right\|_2^2, \label{e_ProtoNet}
\end{align}
where $\mathcal{D}_\text{support, c}$ is the subset of $\mathcal{D}_\text{support}$ that contains examples of class $c$. That is, ProtoNet performs nearest centroid classification and \autoref{e_ProtoNet} aims to learn a good feature extractor. Similar approaches are~\cite{Qiao2018Few,sung2018learning,VinyalsBLKW16Matching,Li2019Finding,Ye2018Learning}.

\subsubsection{Meta-learning with the target classifier}
\label{ss_target_literature}
In \autoref{eq_meta},  
the query set $\mathcal{D}_\text{query}$ is the main supervision for meta-training the few-shot learner $\mathcal{A}$. That is, $\mathcal{D}_\text{query}$ is used to measure the loss of the classifier $\mathcal{A}(\mathcal{D}_\text{support})$ outputted by $\mathcal{A}$. Such a loss is then used to derive gradients to update $\mathcal{A}$.

Instead of using the query set $\mathcal{D}_\text{query}$ to measure the quality of $\mathcal{A}(\mathcal{D}_\text{support})$ in meta-training the few-shot learner $\mathcal{A}$,
\cite{WangH16Learning,WangRH17Learning} constructed the ``target'' classifier $h^\star$, which is the $C$-way classifier
trained with ample labeled examples of the same classes in ${D}_\text{support}$. 
The target classifier $h^\star$ can be seen as the classifier that we hope the few-shot learner $\mathcal{A}$ to output given $\mathcal{D}_\text{support}$: after all, what we ask the few-shot learner to do is to generate a classifier as if it were trained with ample examples. 
To this end, one can measure the loss of $\mathcal{A}(\mathcal{D}_\text{support})$ by its difference from the target classifier $h^\star$ in the parameter space, and use the loss to meta-train the few-shot learner $\mathcal{A}$. We note that computing the difference in the parameter space does not require the query set $\mathcal{D}_\text{query}$. For instance, in \cite{WangH16Learning}, 
a simulated FSL task $\mathcal{T}\in\mathcal{D}_\text{base}$
is formulated as a tuple $(\mathcal{D}_\text{support},h^\star)$, in which the target classifier $h^\star$ is obtained by training with all the data in $\mathcal{D}_\text{base}$ (that are from the same classes in $\mathcal{D}_\text{support}$).
Accordingly, \autoref{eq_meta} for meta-training $\mathcal{A}$ can be reformulated as follows
\begin{align}
    \min_\mathcal{A} \sum_{\mathcal{T}\sim\mathcal{D}_\text{base}} \; \ell_\text{model}(\mathcal{A}(\mathcal{D}_\text{support}), h^\star), \label{eq_meta_target}
\end{align}
where $\ell_\text{model}$ is a distance measure (\eg, Euclidean distance) between classifiers in the parameter space. 

One advantage of \autoref{eq_meta_target} over \autoref{eq_meta}, or a few-shot task $(\mathcal{D}_\text{support},h^\star)$ over a few-shot task $(\mathcal{D}_\text{support},\mathcal{D}_\text{query})$, is the \emph{effective} amount of supervised signals associated with each support set $\mathcal{D}_\text{support}$. 
Unlike $(\mathcal{D}_\text{support},\mathcal{D}_\text{query})$, which is limited by the query set size $|\mathcal{D}_\text{query}|$, $h^\star$ can be trained with all the data in $\mathcal{D}_\text{base}$ (from the same classes in $\mathcal{D}_\text{support}$) and essentially  provides more information for meta-training. 

\begin{figure*}
	\centering
	\begin{minipage}[t]{\linewidth}
		\centering
		\includegraphics[width=\linewidth]{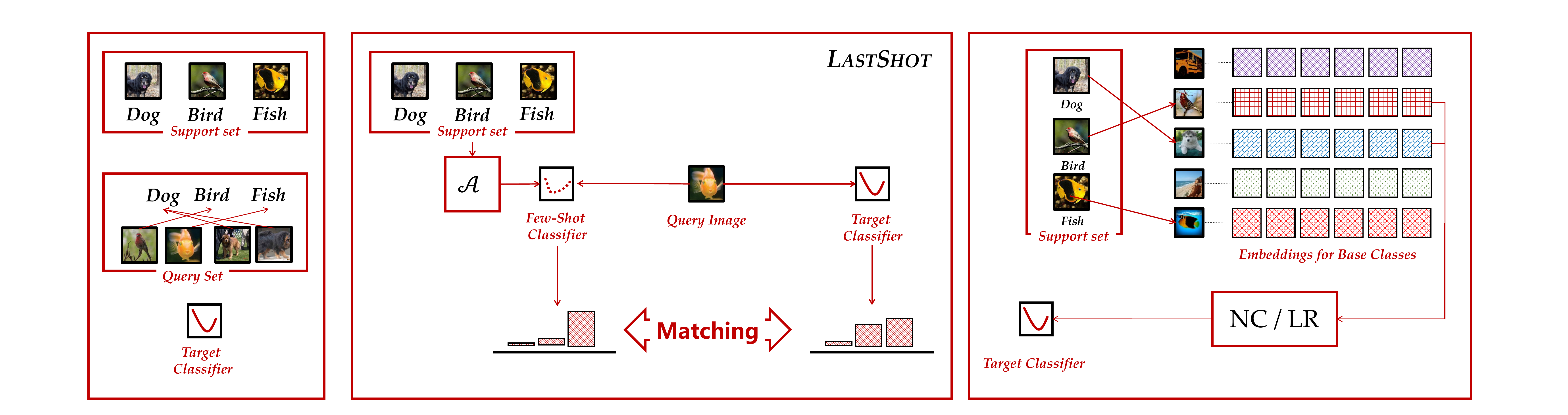}\\
	\end{minipage}
	\caption{An illustration of {\Last} for few-shot classification. {\it Left}: the few-shot task $\mathcal{T}$ in \Last, which contains a support set $\mathcal{D}_\text{support}$, a query set $\mathcal{D}_\text{query}$, and a target classifier $h^\star$. {\it Middle}: the learning objective of \Last, in which the
	few-shot learner $\mathcal{A}(\mathcal{D}_\text{support})(\vx)$ compares its prediction on a query example $\vx$ to the target classifier $h^\star(\vx)$ to receive supervision for updates.
	{\it Right}: our proposed way to construct the $C$-way target classifier $h^\star$ for teaching $\mathcal{A}$. Based on the pre-extracted features of the base class data, \Last can build a strong nearest centroid classifier (NC) or a $C$-way logistic regression (LR) efficiently.}\label{fig:metarepo}
\end{figure*}

Denote by $\vphi$ and $\vphi^\star$ the parameters of $h=\mathcal{A}(\mathcal{D}_\text{support})$ and $h^\star$, \cite{WangH16Learning,WangRH17Learning,Gui2018Few} computed the squared Euclidean distance $\ell_\text{model}(h, h^\star) = \|\vphi-\vphi^\star\|_2^2$ to drive meta-training. Doing so, however, has two drawbacks. First, the Euclidean distance may not be suitable in the high-dimensional parameter space. Indeed, \cite{WangH16Learning,WangRH17Learning} have limited their approaches to only meta-train the last fully-connected layer of a neural network. Second, the Euclidean distance requires the two classifiers to have exactly the same architecture, meaning that we have to re-construct $h^\star$ for different meta-learning algorithms and models.

\subsection{\Last: Learning with a strong teacher}
\label{ss_ours_last}
Let us now introduce our approach. We define each simulated FSL task $\mathcal{T} = (\mathcal{D}_\text{support}, \mathcal{D}_\text{query}, h^\star)\sim\mathcal{D}_\text{base}$ as a tuple that contains both the query set and the target classifier beside the support set. The target classifier $h^\star$ is constructed similarly as described in 
\autoref{ss_target_literature}, \ie, using all the data in $\mathcal{D}_\text{base}$ that are from the same classes as in $\mathcal{D}_\text{support}$. We will provide more details in \autoref{ss_cont_target}.
We propose a new meta-learning objective to meta-train the few-shot learner $\mathcal{A}$ with this new definition of $\mathcal{T}$,

\begin{align}
    \min_\mathcal{A} \sum_{\mathcal{T}\sim\mathcal{D}_\text{base}} \;\sum_{(\vx,y)\in\mathcal{D}_\text{query}} & {\ell_\Last(\mathcal{A}(\mathcal{D}_\text{support})(\vx), h^\star(\vx))} \nonumber \\ 
    & + \lambda\times\ell(\mathcal{A}(\mathcal{D}_\text{support})(\vx), y). \label{eq_Last}
\end{align}
This objective is as general as \autoref{eq_meta} but is able to benefit from the target classifiers $h^\star$.
Specifically, \autoref{eq_Last} involves two loss terms on the query data. One is exactly the same as in \autoref{eq_meta}, comparing $\mathcal{A}(\mathcal{D}_\text{support})(\vx)$ to $y$. The other term compares $\mathcal{A}(\mathcal{D}_\text{support})(\vx)$ to $h^\star(\vx)$ via a loss function $\ell_\Last$, which can be interpreted as a relaxed version of \autoref{eq_meta_target}: \emph{to match the predictions of the two classifiers, not their parameters.} This relaxed version effectively overcomes the two drawbacks of \autoref{eq_meta_target}. On the one hand, we are now able to employ a stronger target classifier (\eg, the entire neural network model) to teach the few-shot learner, since the comparison between them is in the lower-dimensional space govern by the number of ways $C$, not by the number of parameters of $h^\star$. On the other hand, comparing the predictions allows us to disentangle the model architectures of the few-shot learner, its generated classifier, and $h^\star$. In other words, the same $h^\star$ associated with each $\mathcal{D}_\text{support}$ can now be applied to train different kinds of few-shot learners, making \autoref{eq_Last} a widely applicable objective for meta-learning. Indeed, wherever \autoref{eq_meta} can be applied, \autoref{eq_Last} should be equally applicable.

We realize the loss $\ell_\Last$ in \autoref{eq_Last} by the 
Kullback–Leibler (KL) divergence, inspired by \cite{chen2020big}. That is, instead of comparing the hard one-hot predictions, we compare the soft predictions that encode the target classifier's confidence as well as the learned semantic relationship among classes.
More specifically, we take the logits $\vpsi(\vx)\in\R^C$ and $\vpsi^\star(\vx)\in\R^C$ produced by $h=\mathcal{A}(\mathcal{D}_\text{support})$ and $h^\star$ on a query example $\vx$, smooth the latter with a temperature $\tau > 0$ \cite{lopez2016unifying}, and normalize them with the $\mathbf{softmax}$ function before computing the KL divergence, 
\begin{align}
& \ell_\Last(\mathcal{A}(\mathcal{D}_\text{support})(\vx), h^\star(\vx)) = \nonumber \\
& \mathbf{KL}\left( \mathbf{softmax}(\frac{\vpsi^\star(\vx)}{\tau}) \;\big{\|}\; \mathbf{softmax}\left(\vpsi(\vx)\right)\right)\;. \label{eq_KL}
\end{align}
We note that the KL divergence is not the only way to realize $\ell_\Last$. One can apply other measures that can characterize the difference of predictions between two classifiers.

\subsection{Construction of the target classifier}
\label{ss_cont_target}
So far, we assume that the target classifier $h^\star$ is well-specified and can be easily prepared for each sampled FSL task $\mathcal{T}$. 
In this subsection,
we investigate several ways to construct it. We note that, a target classifier $h^\star$ for each $\mathcal{T}$ should be a $C$-way classifier; the $C$ classes are the same as those contained in $\mathcal{D}_\text{support}$ and $ \mathcal{D}_\text{query}$. That is, $h^\star$ would be different across the simulated FSL tasks.

\subsubsection{Masking the $B$-way base classifier}
Given the $B$-class many-shot training set $\mathcal{D}_\text{base}$, one intuitive way to construct $h^\star$ is to directly train a $B$-class classifier. However, since most of the FSL settings have $C$ (\ie, the number of ways in an FSL task $\mathcal{T}$) much smaller than $B$, we must mask out the logits of $h^\star$ not from the $C$ classes in $\mathcal{T}$, followed by calibration. In our
experiments, we found such a mismatch in class numbers does degrade the performance of \Last (please see \autoref{sec:ablation}). We thus develop two simple, efficient, yet effective alternatives to construct $h^\star$.

\subsubsection{Nearest centroid classifiers (NC)}
Let us denote by $f^\ddagger$ the feature extractor of the learned $B$-class classifier and by $\mathcal{C}_\mathcal{T}$ the class label space of $\mathcal{T}$. Our first idea is to build a $C$-way nearest centroid classifier \cite{Wang2019SimpleShot,mensink2013distance}. Denote by $\mathcal{D}_\text{base, c}$ the subset of $\mathcal{D}_\text{base}$ of class $c$, we can pre-compute the class mean $\vmu_c$ for each class $c$ 
\begin{align}
\vmu_c = \sum_{\vx'\in\mathcal{D}_\text{base, c}}  \frac{f^\ddagger(\vx')}{|\mathcal{D}_\text{base, c}|}.
\end{align} 
Then, given a $C$-way few-shot task $\mathcal{T}$, we can retrieve $\vmu_c$ for class $c\in\mathcal{C}_\mathcal{T}$ and compute for $\vx\in\mathcal{D}_\text{query}$ the logit of class $c$
\begin{align}
	\psi^{\star}_c(\vx) = -\left\|f^\ddagger(\vx) - \vmu_c\right\|_2^2.
\end{align}
That is, the logit vector $\vpsi^\star(\vx)\in\R^C$ of the target classifier for $\vx$ is the negative squared distance between $f^\ddagger(\vx)$ and the $C$ class means, which is then used in \autoref{eq_KL}.
We note that all the $f^\ddagger(\vx), \forall \vx\in\mathcal{D}_\text{base}$ can be pre-extracted. 

\subsubsection{Logistic regression (LR)}
Our second way is to train a $C$-class logistic regression using the features pre-extracted by $f^\ddagger$. Specifically, given a $C$-way few-shot task $\mathcal{T}$, we gather the pre-extracted features $f^\ddagger(\vx)$ for data in the subset $\cup_{c\in\mathcal{C}_\mathcal{T}} \mathcal{D}_\text{base, c}$ of $\mathcal{D}_\text{base}$ and train a $C$-class logistic regression. 
The logit vector $\vpsi^\star(\vx)$ of the target classifier for $\vx$ is the output of the LR model. Training the linear LR is efficient given the extracted features.

\subsection{Querying the target classifier in meta-training}
Besides different ways of constructing $h^\star$, we also investigate how to query $h^\star$; \ie, how to obtain $\vpsi^\star(\vx)$ for \autoref{eq_KL}. The basic way is to input the query example $\vx$ directly. Motivated by recent studies \cite{xie2020self,radosavovic2018data}, where strengthening the teacher or weakening the student (\eg, adding noise) could further improve the distillation or self-training performance, we explore (a) varying the size of the input image for $\vx$ or (b) applying autoaugment \cite{cubuk2019autoaugment} to $\vx$. We note that both operations are applied only in the meta-training phase.

\subsubsection{Varying the input image size} 
One way to strengthen the teacher target classifier $h^\star$, \ie, to increase its prediction accuracy, is to operate on images with a larger size. 
Specifically, when we apply $f^\ddagger$ to extract features to construct $h^\star$, we enlarge the input images (\eg, by $140\%$). We do so also when we query $h^\star$; \ie, we enlarge $\vx$ before applying $h^\star(\vx)$. 
Namely, the target classifier is constructed with and applied to images of a larger size, resulting in better logits $\vpsi^\star(\vx)$ to be matched in \autoref{eq_KL}.

\emph{We note that, we enlarge the image size only for constructing and querying the teacher classifier during meta-training. We do not change the image size for the few-shot learner and its generated classifier. Therefore, our method does not raise any concern of unfair comparisons to existing methods.}

\subsubsection{Applying autoaugment~\cite{cubuk2019autoaugment}}
We further consider weakening the student model (\ie, the few-shot learner and its generated classifier) to encourage it to learn more from the teacher. 
Specifically, we apply autoaugment~\cite{cubuk2019autoaugment} to each query example $\vx$, so that both the teacher $h^\star$ and the student $h=\mathcal{A}(\mathcal{D}_\text{support})$ make their predictions on a noisy query example. Since the teacher is trained on a larger data set $\mathcal{D}_\text{base}$, it suffers less than the student from the noise, creating a larger training signal in \autoref{eq_KL}.

We find that both treatments can improve \Last on some of the datasets, and strengthening the teacher is more stable.
Thus, we mainly consider it in the experiments. 

\subsection{Evaluation of the learned few-shot learner $\mathcal{A}$}

So far in \autoref{sec:approach}, we focus on how to meta-train the few-shot learner $\mathcal{A}$ that can output a classifier $h=\mathcal{A}(\mathcal{D}_\text{support})$ given the support set $\mathcal{D}_\text{support}$ of a few-shot task $\mathcal{T}$. The few-shot task $\mathcal{T}$ during meta-training is sampled/simulated from ${D}_\text{base}$, and one can flexibly define it as $\mathcal{T}=(\mathcal{D}_\text{support}, \mathcal{D}_\text{query})$, $\mathcal{T}=(\mathcal{D}_\text{support}, h^\star)$, or $\mathcal{T}=(\mathcal{D}_\text{support}, \mathcal{D}_\text{query}, h^\star)$ (cf. \autoref{ss_basic_meta}, \autoref{ss_target_literature}, \autoref{ss_ours_last}, respectively).

During meta-testing, in which the trained few-shot learner $\mathcal{A}$ is to be evaluated on novel classes (cf. \autoref{ss_notation}), we follow the standard evaluation protocol (cf. \autoref{ss_protocal}): each meta-testing task $\mathcal{T}'$ is a tuple $(\mathcal{D}_\text{support}', \mathcal{D}_\text{query}')$. The accuracy on a task is defined as
\begin{align}
    \frac{1}{|\mathcal{D}_\text{query}'|}\sum_{(\vx,y)\in\mathcal{D}_\text{query}'} \textbf{1}[\mathcal{A}(\mathcal{D}_\text{support}')(\vx) == y],
\end{align}
where $\textbf{1}$ is an indicator function whose value is 1 if the input is true; otherwise, 0. In other word, during meta-testing, we do not need a target classifier for each meta-testing task. We construct the target classifiers only to facilitate meta-training, and this is possible because the base class data ${D}_\text{base}$ contains sufficient examples.
\section{Experiments on Few-Shot Classification}
\label{sec:experiment}
\begin{table}[tbp]	
\centering
\caption{Statistics of the five benchmark datasets. We list the number of classes and number of images in each split.}
	\tabcolsep 2pt
	\begin{tabular}{l|cc|cc|cc}
		\addlinespace
		\toprule
	     & \multicolumn{2}{c|}{\bf meta-train} & \multicolumn{2}{c|}{\bf meta-val} & \multicolumn{2}{c}{\bf meta-test} \\
		\midrule
		dataset & Class   & Image  & Class   & Image  & Class  & Image \\
		\midrule
		{\it mini}ImageNet~\cite{VinyalsBLKW16Matching} & 64 &  38,400 & 16 & 9,600 & 20 & 12,000 \\
		{\it tiered}ImageNet~\cite{Ren2018Meta} & 351 & 448,695 & 97 & 124,261 & 160 &  206,209 \\
		CUB~\cite{WahCUB_200_2011} & 100 & 5,891 & 50 & 2,932 & 50 & 2,965 \\
		CIFAR-FS~\cite{bertinetto2018meta} & 64 &  38,400 & 16 & 9,600 & 20 & 12,000 \\
		FC-100~\cite{Oreshkin2018TADAM} & 60 & 36,000 & 20 & 12,000 & 20 & 12,000 \\
		\bottomrule
	\end{tabular}
	\label{tab:dataset}
\end{table}

\subsection{Datasets}
We evaluate our approach \Last using multiple benchmark datasets, including \textbf{\emph{mini}ImageNet} \cite{VinyalsBLKW16Matching}, \textbf{\emph{tiered}ImageNet} \cite{Ren2018Meta}, \textbf{CUB} \cite{WahCUB_200_2011}, \textbf{CIFAR-FS} \cite{bertinetto2018meta}, and \textbf{FC-100} \cite{Oreshkin2018TADAM}. There are 100 classes in {\it mini}ImageNet with 600 images per class. Following the split of~\cite{ravi2017optimization}, the meta-training/validation/testing sets contain 64/16/20 (non-overlapping) classes, respectively. In {\it tiered}ImageNet~\cite{Ren2018Meta}, the class numbers in the three sets are 351/97/160, respectively, again without overlapping.
For CUB, we follow the split of~\cite{Ye2018Learning} and set 100 classes for meta-training. Two 50-class splits from the remaining classes are used for meta-validation and meta-testing. 
CIFAR-FS~\cite{bertinetto2018meta} and FC-100~\cite{Oreshkin2018TADAM} are two variants of the CIFAR-100 dataset~\cite{krizhevsky2009learning} for few-shot learning but with different partitions. The classes in CIFAR-FS are randomly split into 64/16/20 for meta-training/validation/testing, while FC-100 takes the split of 60/20/20 classes following the super-classes. There are 600 images in each class in CIFAR-100.
All images are resized to 84 by 84 for the first three datasets and are resized to 32 by 32 for the latter two CIFAR variants in advance.
The detailed statistics are listed in \autoref{tab:dataset}. 

\subsection{Evaluation protocols} 
\label{ss_protocal}
We follow the evaluation in the literature~\cite{rusu2019meta,Ye2018Learning,Chao2019Meta,Zhang2020Deep}, where 10,000 $C$-way $K$-shot tasks are sampled from the meta-testing set. In each task, the query set contains 15 instances in each class. Mean accuracy (in \%) as well as the 95\% confidence interval are reported. For every meta-training task, we also sample 15 instances per class to construct the query set, unless stated otherwise.

\subsection{Implementation details}
\Last is a plug-and-play approach that is compatible with most of the existing meta-learning FSL methods.
We apply \Last to four representative meta-learning methods: ProtoNet~\cite{SnellSZ17Prototypical}, FEAT~\cite{Ye2018Learning}, ProtoMAML~\cite{triantafillou2019meta}, and MetaOptNet~\cite{Lee2019Meta}. The first two are embedding-based methods to learn good feature extractors; the latter two are similar to MAML that learn in a two-stage inner/outer-loop manner.

\subsubsection{Model architectures for few-shot classification}
We follow \cite{Lee2019Meta} to use a ResNet-12 \cite{he2016deep} for the feature extractor $f_{\vtheta}$ (cf. \autoref{ss_meta-examplar}), which contains a wider width and the Dropblock module~\cite{Ghiasi2018Drop} to avoid over-fitting.

\subsubsection{Initialization of the feature extractor}
Following~\cite{Wang2019SimpleShot,Ye2018Learning}, we initialize the feature extractor $f_{\vtheta}$ used in our few-shot learner and our target classifier using the base class data $\mathcal{D}_\text{base}$. Concretely, we learn $f_{\vtheta}$, together with a 
$B$-way linear classifier $W$ on top\footnote{We omit the bias term for brevity.}, to classify the base classes. For instance, on {\it mini}ImageNet, $B$ is $64$, meaning that we pre-train a $64$-class classifier.
More specifically, we train the classifier by minimizing the following objective over examples $(\vx_i, y_i)$ in $\mathcal{D}_\text{base}$:
\begin{align}
	\min_{W, f_{\vtheta}}\; \sum_{(\vx_i, y_i)\in{\mathcal{D}_\text{base}}} \; \ell(W^\top f_{\vtheta}(\vx_i), \;y_i).\label{eq:cls}
\end{align}
Here, $\ell(\cdot, \cdot)$ measures the discrepancy between the predicted label and the ground-truth label, and we use the cross-entropy loss. We follow~\cite{Ye2018Learning} for optimization. For instance, on {\it mini}ImageNet, \autoref{eq:cls} is optimized by stochastic gradient descent (SGD) with momentum 0.9, weight decay 0.0005, batch-size 128, and initial learning rate 0.1. After each epoch, we validate the quality of $f_{\vtheta}$ on the meta-validation set by sampling 200 1-shot $C$-way tasks where $C$ equals the number of classes in the meta-validation set (\eg, 16 for {\it mini}ImageNet). For each task, we use the ProtoNet decision rule in \autoref{e_ProtoNet} to evaluate the classification performance. Similar procedures are employed for other datasets.

With the pre-trained $f_{\vtheta}$, we can initialize the few-shot learner in the meta-training phase. For fair comparisons, we use the notation $\dagger$ to denote our re-implemented results of the baseline methods using this pre-training procedure, which usually leads to higher accuracy than the reported results. We note that, FEAT~\cite{Ye2018Learning} already uses the above procedure, so we compare with the published results directly.

\subsubsection{Meta-Learning with \Last}
In our implementation, we do not use the meta-validation set to learn the few-shot learner; \ie, the meta-validation set is only used to select the best hyper-parameters such as the balance coefficient $\lambda$ in \autoref{eq_Last} or the temperature $\tau$. For \Last with LR, we randomly select at most 50 instances per class to construct a linear logistic regression for each few-shot task. To strengthen the teacher, we enlarge 
images during training and querying the target classifier $h^\star$.
For {\it mini}ImageNet, {\it tiered}ImageNet, and CUB, we enlarge the images to 116 by 116. 
For CIFAR-FS and FC-100, we enlarge the images to 50 by 50.
\emph{We note that, we do not enlarge the images for the few-shot learner and its outputted classifier. Thus, our results can be fairly compared to the literature.}

\subsubsection{Non-meta-learning based FSL methods}
The initialized $f_{\vtheta}$ (cf. \autoref{eq:cls}) can be used for few-shot classification directly. For instance, by extracting features on the support set data of the novel classes using $f_{\vtheta}$, we can apply \autoref{e_ProtoNet} to classify a query example according to the nearest centroid classification rule. We call this method PT-EMB, which stands for pre-trained-embedding. We also compare to SimpleShot\cite{Wang2019SimpleShot}, which employs several simple pre-processing steps to further improve the accuracy.

\begin{table}[t]
	\centering
	\newcolumntype{g}{>{\columncolor{LightCyan}}c}
	\newcolumntype{f}{>{\columncolor{LightCyan}}l}
	\caption{Few-shot classification accuracy plus 95\% confidence interval on {\it mini}ImageNet with the ResNet-12 backbone. Results are evaluated over 10,000 tasks. $\dagger$ denotes our re-implemented results. Note that the cited results of FEAT~\cite{Ye2018Learning} are implemented with the same pre-training procedure already.}
	\tabcolsep 3pt
	%{\small\begin{tabular}{c|gg}
	\begin{tabular}{c|cc}
			%{\small\begin{tabular}{@{\;}c@{\;}|@{\;}cc@{\;}}
			\addlinespace
			\toprule
			Setting & 1-Shot     & 5-Shot \\
			\midrule
            ModelRegression~\cite{WangH16Learning} & 61.94  $\pm$  0.20 & 76.24  $\pm$  0.14\\			MetaOptNet{~\cite{Lee2019Meta}} & 62.64  $\pm$  0.20 & 78.63  $\pm$  0.14 \\
			SimpleShot$^\dagger${~\cite{Wang2019SimpleShot}} & 65.36  $\pm$  0.20 & 81.39  $\pm$  0.14 \\
			TRAML+ProtoNet{~\cite{Li2020Boosting}} & 60.31  $\pm$  0.48 & 77.94  $\pm$  0.57 \\
			RFS-Simple{~\cite{Tian2929Rethinking}} & 62.02  $\pm$  0.63 & 79.64  $\pm$  0.44 \\
			RFS-Distill{~\cite{Tian2929Rethinking}} & 64.82  $\pm$  0.60 & 82.14  $\pm$  0.43 \\
			DSN-MR{~\cite{Simon2020Adaptive}} & 64.60  $\pm$  0.72 & 79.51  $\pm$  0.50 \\
			MTL+E3BM{~\cite{Liu2020Ensemble}} & 63.80  $\pm$  0.40 & 80.10  $\pm$  0.30 \\
			DeepEMD{~\cite{Zhang2020Deep}} & 65.91  $\pm$  0.82 & 82.41  $\pm$  0.56 \\
			TRAML+AM3{~\cite{Li2020Boosting}} & 67.10  $\pm$  0.54 & 79.54  $\pm$  0.60 \\
			\midrule\midrule
			ProtoNet$^\dagger${~\cite{SnellSZ17Prototypical}} & 62.39  $\pm$  0.20 & 79.74  $\pm$  0.14 \\
			\rowcolor{LightCyan}
			+ \Last (NC) & 64.16  $\pm$  0.20 & 81.23  $\pm$  0.14 \\
			\rowcolor{LightCyan}
			+ \Last (LR) & 64.80  $\pm$  0.20 & 81.65  $\pm$  0.14 \\
			\midrule
			ProtoMAML$^\dagger${~\cite{triantafillou2019meta}} & 62.04  $\pm$  0.21 & 79.62  $\pm$  0.14 \\
			\rowcolor{LightCyan}
			+ \Last (NC) & 63.07  $\pm$  0.20 & 81.04  $\pm$  0.14 \\
			\rowcolor{LightCyan}
			+ \Last (LR) & 63.05  $\pm$  0.20 & 81.11  $\pm$  0.14 \\
			\midrule
			MetaOptNet$^\dagger${~\cite{Lee2019Meta}} & 63.21  $\pm$  0.20 & 79.94  $\pm$  0.14 \\
			\rowcolor{LightCyan}
			+ \Last (NC) & 65.07  $\pm$  0.20 & 80.34  $\pm$  0.14 \\
			\rowcolor{LightCyan}
			+ \Last (LR) & 65.08  $\pm$  0.20 & 80.40  $\pm$  0.14 \\
			\midrule
			FEAT{~\cite{Ye2018Learning}}  & 66.78  $\pm$  0.20 & 82.05  $\pm$  0.14 \\
			\rowcolor{LightCyan}
			+ \Last (NC) & {67.33  $\pm$  0.20} & 82.39  $\pm$  0.14 \\
			\rowcolor{LightCyan}
			+ \Last (LR) &\bf 67.35  $\pm$  0.20 & {\bf 82.58  $\pm$  0.14} \\
			\bottomrule
	\end{tabular}
	\label{tab:miniimagenet}
\end{table}

\begin{table}[t]
	\centering
	\newcolumntype{g}{>{\columncolor{LightCyan}}c}
	\newcolumntype{f}{>{\columncolor{LightCyan}}l}
	\caption{Few-shot classification accuracy plus 95\% confidence interval on {\it tiered}ImageNet with the ResNet-12 backbone. Results are evaluated over 10,000 tasks. $\dagger$ denotes our re-implemented results. Note that the cited results of FEAT~\cite{Ye2018Learning} are implemented with the same pre-training procedure already.}
	\tabcolsep 3pt
		\begin{tabular}{@{\;}c@{\;}|cc}
			\addlinespace
			\toprule
			Setting & 1-Shot     & 5-Shot \\
			\midrule
			MetaOptNet{~\cite{Lee2019Meta}} & 65.99  $\pm$  0.72 & 81.56  $\pm$  0.53 \\
			SimpleShot$^\dagger${~\cite{Wang2019SimpleShot}} & 70.51  $\pm$  0.23 & 84.58  $\pm$  0.16 \\
			RFS-Simple{~\cite{Tian2929Rethinking}} & 69.74  $\pm$  0.72 & 84.41  $\pm$  0.55 \\
			DSN{~\cite{Simon2020Adaptive}}   & 66.22  $\pm$  0.75 & 82.79  $\pm$  0.48 \\
			DSN-MR{~\cite{Simon2020Adaptive}} & 67.39  $\pm$  0.82 & 82.85  $\pm$  0.56 \\
			RFS-Distill{~\cite{Tian2929Rethinking}} & 71.52  $\pm$  0.69 & \bf 86.03  $\pm$  0.49 \\
			MTL+E3BM{~\cite{Liu2020Ensemble}} & 71.20  $\pm$  0.40 & 85.30  $\pm$  0.30 \\
			DeepEMD{~\cite{Zhang2020Deep}} & 71.16  $\pm$  0.87 & {\bf 86.03  $\pm$  0.58} \\
			\midrule\midrule
			ProtoNet$^\dagger${~\cite{SnellSZ17Prototypical}} & 68.23  $\pm$  0.23 & 84.03  $\pm$  0.16 \\
			\rowcolor{LightCyan}
			+ \Last (NC) & 68.91  $\pm$  0.23 & 85.15  $\pm$  0.16 \\
			\rowcolor{LightCyan}
			+ \Last (LR) & 69.37  $\pm$  0.23 & 85.36  $\pm$  0.16 \\
			\midrule
			ProtoMAML$^\dagger${~\cite{triantafillou2019meta}} & 67.10  $\pm$  0.23 & 81.18  $\pm$  0.16 \\
			\rowcolor{LightCyan}
			+ \Last (NC) & 68.42  $\pm$  0.23 & 83.63  $\pm$  0.16 \\
			\rowcolor{LightCyan}
			+ \Last (LR) & 68.80  $\pm$  0.23 & 83.72  $\pm$  0.16 \\
			\midrule
			MetaOptNet$^\dagger${~\cite{Lee2019Meta}} & 66.49  $\pm$  0.23 & 82.36  $\pm$  0.16 \\
			\rowcolor{LightCyan}
			+ \Last (NC) & 68.23  $\pm$  0.23 & 83.22  $\pm$  0.16 \\
			\rowcolor{LightCyan}
			+ \Last (LR) & 68.29  $\pm$  0.23 & 83.45  $\pm$  0.16 \\
			\midrule
			FEAT{~\cite{Ye2018Learning}}  & 71.13  $\pm$  0.23 & 84.79  $\pm$  0.16 \\
			\rowcolor{LightCyan}
			+ \Last (NC) & 72.03  $\pm$  0.23 & 85.66  $\pm$  0.16 \\
			\rowcolor{LightCyan}
			+ \Last (LR) & {\bf 72.43  $\pm$  0.23} & 85.82  $\pm$  0.16 \\
			\bottomrule
	\end{tabular}
	\label{tab:tieredimagenet}
\end{table}

\begin{table}[t]
	\centering
	\newcolumntype{g}{>{\columncolor{LightCyan}}c}
	\newcolumntype{f}{>{\columncolor{LightCyan}}l}
	\caption{Few-shot classification accuracy plus 95\% confidence interval on CUB with the ResNet-12 backbone. Results are evaluated over 10,000 tasks. $\dagger$ denotes our re-implemented results. Note that the cited results of FEAT~\cite{Ye2018Learning} are implemented with the same pre-training procedure already.}
	\tabcolsep 4pt
		\begin{tabular}{c|cc}
			\addlinespace
			\toprule
			Setting & 1-Shot     & 5-Shot \\
			\midrule
			TriNet{~\cite{Chen2019Multi}} & 69.61  $\pm$  0.46 & 84.10  $\pm$  0.35 \\
			DeepEMD{~\cite{Zhang2020Deep}} & 75.65  $\pm$  0.83 & 88.69  $\pm$  0.50 \\
			Negative-Cosine{~\cite{Liu2020Negative}} & 72.66  $\pm$  0.85 & 89.40  $\pm$  0.43 \\
			\midrule\midrule
			ProtoNet{~\cite{SnellSZ17Prototypical}} & 72.92  $\pm$  0.21 & 87.84  $\pm$  0.12 \\
			\rowcolor{LightCyan}
			+ \Last (NC) & 75.83  $\pm$  0.21 & 89.83  $\pm$  0.12 \\
			\rowcolor{LightCyan}
			+ \Last (LR) & 75.80  $\pm$  0.21 & 90.22  $\pm$  0.12 \\
			\midrule
			ProtoMAML{~\cite{triantafillou2019meta}} & 71.07  $\pm$  0.21 & 87.27  $\pm$  0.12 \\
			\rowcolor{LightCyan}
			+ \Last (NC) & 74.57  $\pm$  0.21 & 88.67  $\pm$  0.12 \\
			\rowcolor{LightCyan}
			+ \Last (LR) & 74.18  $\pm$  0.21 & 88.23  $\pm$  0.12 \\
			\midrule
			MetaOptNet{~\cite{Lee2019Meta}} & 75.31  $\pm$  0.21 & 89.24  $\pm$  0.12 \\
			\rowcolor{LightCyan}
			+ \Last (NC) & 78.54  $\pm$  0.21 & 90.46  $\pm$  0.12 \\
			\rowcolor{LightCyan}
			+ \Last (LR) & 78.79  $\pm$  0.21 & 90.46  $\pm$  0.12 \\
			\midrule
			FEAT{~\cite{Ye2018Learning}}  & 77.97  $\pm$  0.21 & 90.03  $\pm$  0.12 \\
			\rowcolor{LightCyan}
			+ \Last (NC) & 80.07  $\pm$  0.21 & 91.43  $\pm$  0.12 \\
			\rowcolor{LightCyan}
			+ \Last (LR) & {\bf 80.20  $\pm$  0.21} & {\bf 91.49  $\pm$  0.12} \\
			\bottomrule
	\end{tabular}
	\label{tab:cub}
\end{table}

\begin{table*}[tbp]
	\centering
	\newcolumntype{g}{>{\columncolor{LightCyan}}c}
	\newcolumntype{f}{>{\columncolor{LightCyan}}l}
	\caption{
		Few-shot classification accuracy plus plus 95\% confidence interval on two splits of CIFAR-100, \ie, CIFAR-FS and FC-100, based on the ResNet-12 backbone. 
		We use ``N/A'' if the confidence interval is not reported in the cited papers. $\dagger$ denotes our re-implemented results. Note that the cited results of FEAT~\cite{Ye2018Learning} are implemented with the same pre-training procedure already. $\mathsection$: DeepEmd~\cite{Zhang2020Deep} make predictions for FC-100 by meta-training and meta-testing on larger images with size 84$\times$ 84 instead of 32$\times$32. }
	\begin{tabular}{l|cc||cc}
		\addlinespace
		\toprule
		Dataset & \multicolumn{2}{c||}{\bf CIFAR-FS} & \multicolumn{2}{c}{\bf FC-100} \\
		Setting & 1-Shot 5-Way & 5-Shot 5-Way & 1-Shot 5-Way & 5-Shot 5-Way\\
		\midrule
		ProtoNet{~\cite{SnellSZ17Prototypical}}    & 72.20  $\pm$  0.70 & 83.50  $\pm$  0.50 & 37.50  $\pm$  0.60 & 52.50  $\pm$  0.60 \\
		ShotFree{~\cite{Ravichandran2019Few}}    & 69.20  $\pm$  N/A & 84.70  $\pm$  N/A & - & - \\
		MetaOptNet{~\cite{Lee2019Meta}}           & 72.00  $\pm$  0.70 & 84.20  $\pm$  0.50 & 41.10  $\pm$  0.60 & 55.50  $\pm$  0.60 \\
		RFS-Simple{~\cite{Tian2929Rethinking}}       & 71.50  $\pm$  0.80 & 86.00  $\pm$  0.50 & 42.60  $\pm$  0.70 & 58.10  $\pm$  0.60 \\
		RFS-Distill{~\cite{Tian2929Rethinking}}       & 73.90  $\pm$  0.80 & 86.90  $\pm$  0.50 & 44.60  $\pm$  0.70 & 60.90  $\pm$  0.60 \\
		DeepEmd$^\mathsection${~\cite{Zhang2020Deep}}       & - & - & \bf 46.47  $\pm$  0.78 & \bf 63.22  $\pm$  0.71 \\
		\midrule\midrule
		ProtoNet$^\dagger${~\cite{SnellSZ17Prototypical}} & 72.42  $\pm$  0.21 & 85.26  $\pm$  0.12  & 42.01  $\pm$  0.18 & 57.48  $\pm$  0.18 \\
		\rowcolor{LightCyan}
		+ \Last (NC) &
		74.77  $\pm$  0.21 & 86.46  $\pm$  0.12  & 42.65  $\pm$  0.18 & 58.43  $\pm$  0.18 \\
		\rowcolor{LightCyan}
		+ \Last (LR) &
		74.69  $\pm$  0.21 & 86.51  $\pm$  0.12  & 42.83  $\pm$  0.18 & 58.16  $\pm$  0.18 \\
		\midrule
		ProtoMAML$^\dagger${~\cite{triantafillou2019meta}} & 69.95  $\pm$  0.21 & 84.61  $\pm$  0.12  & 41.39  $\pm$  0.18 & 56.76  $\pm$  0.18 \\
		\rowcolor{LightCyan}
		+ \Last (NC) & 
		71.90  $\pm$  0.21 & 85.60  $\pm$  0.12  & 41.94  $\pm$  0.18 & 58.26  $\pm$  0.18 \\
		\rowcolor{LightCyan}
		+ \Last (LR) & 
		71.99  $\pm$  0.21 & 85.61  $\pm$  0.12  & 42.23  $\pm$  0.18 & 57.56  $\pm$  0.18 \\
		\midrule
		MetaOptNet$^\dagger${~\cite{Lee2019Meta}} & 72.96  $\pm$  0.22 & 85.50  $\pm$  0.15  & 42.53  $\pm$  0.18 & 57.54  $\pm$  0.18 \\
		\rowcolor{LightCyan}
		+ \Last (NC) &  
		75.66  $\pm$  0.22 & 86.50  $\pm$  0.15  & 43.46  $\pm$  0.18 & 58.30  $\pm$  0.18 \\
		\rowcolor{LightCyan}
		+ \Last (LR) &  
		76.08  $\pm$  0.22 & 86.70  $\pm$  0.15  & 43.29  $\pm$  0.18 & 58.14  $\pm$  0.18 \\
		\midrule
		FEAT$^\dagger${~\cite{Ye2018Learning}}  & 
		75.44  $\pm$  0.21 & 86.56  $\pm$  0.12  & 42.56  $\pm$  0.18 & 57.48  $\pm$  0.18 \\
		\rowcolor{LightCyan}
		+ \Last (NC) & 
		76.67  $\pm$  0.21 & \bf 87.78  $\pm$  0.12  & 43.91  $\pm$  0.18 & 58.41  $\pm$  0.18 \\
		\rowcolor{LightCyan}
		+ \Last (LR) & 
		\bf 76.76  $\pm$  0.21 & 87.49  $\pm$  0.12  & 44.08  $\pm$  0.18 & 59.14  $\pm$  0.18 \\
		\bottomrule
	\end{tabular}
	\label{tab:cifar}
\end{table*}

%%%%%%%%%%%%%%%%%%%%%%%%%%%%

\subsection{Benchmark comparisons}
We first evaluate \Last with 5-Way 1-Shot and 5-Way  5-Shot tasks on {\it mini}ImageNet (\autoref{tab:miniimagenet}), {\it tiered}ImageNet (\autoref{tab:tieredimagenet}), CUB (\autoref{tab:cub}), {CIFAR-FS} (\autoref{tab:cifar}), and {FC-100} (\autoref{tab:cifar}). We highlight the results when a baseline method is combined with \Last via a cyan background in the tables, and make the best performance in each setting in bold. We denote the two implementations of the target classifiers, \ie, the nearest centroid classifier and the logistic regression, by NC and LR, respectively. 
 
We find that \Last can consistently improve each baseline method on all the datasets by 1$\sim$3\%. For example, with \Last, the 1-Shot classification accuracy of FEAT improves from 77.97\% to 80.20\% in \autoref{tab:cub}. By further investigating the four baseline methods, we find that MetaOptNet and FEAT are stronger than ProtoNet and ProtoMAML. In general, \Last can lead to a larger boost for the relatively weaker baselines, but it can still improve MetaOptNet and FEAT.
The target classifiers based on the nearest centroid and LR have similar performance.
Overall, {\Last} variants achieve the state-of-the-art performance on {\it mini}ImageNet and CUB, while obtaining competitive results on {\it tiered}ImageNet. These verify that a strong teacher can provide better supervision in meta-learning.

Last but not the least, comparing to Model Regression \cite{WangH16Learning} and RFS \cite{Tian2929Rethinking}, in which the former only uses the target classifier to guide the few-shot learner in the last fully-connected layer while the latter uses distillation to strengthen the pre-trained features, our \Last outperforms them on {\it mini}ImageNet, CUB, and CIFAR-FS, and performs on par with them on other datasets. The promising performance further demonstrates the effectiveness of our approach.

\begin{table*}[t]
	\centering
	\caption{5-Way $K$-shot classification accuracy and 95\% confidence interval on {\it mini}ImageNet, where $K\in\{1,5,10,20,30,50\}$. We apply \Last on ProtoNet and FEAT, and achieve stable improvements when the shot number becomes larger. For ``PT-EMB'',  we apply the ProtoNet classification rule with the pre-trained feature extractor.}
	\begin{tabular}{c|cccccc}
		\addlinespace
		\toprule
		Setups $\rightarrow$ & 1     & 5     & 10    & 20    & 30    & 50 \\
		\midrule
		PT-EMB & 59.27  $\pm$  0.20 & 80.55  $\pm$  0.14 & 84.37  $\pm$  0.12 & 86.40  $\pm$  0.11 & 87.15  $\pm$  0.10 & 87.74  $\pm$  0.10 \\
		SimpleShot~\cite{Wang2019SimpleShot} & 65.36  $\pm$  0.20 & 81.39  $\pm$  0.14 & 84.89  $\pm$  0.11 & 86.91  $\pm$  0.10 & 87.53 $\pm$ 0.10 & 88.08 $\pm$ 0.10 \\
		\midrule
		ProtoNet~\cite{SnellSZ17Prototypical} & 63.73 $\pm$ 0.21 & 79.40 $\pm$ 0.14 & 82.83 $\pm$ 0.12 & 84.61 $\pm$ 0.11 & 85.07 $\pm$ 0.11 & 85.57 $\pm$ 0.10 \\
		\rowcolor{LightCyan}
		+ \Last (NC) & 64.76 $\pm$ 0.20 & 81.60 $\pm$ 0.14 & 85.03 $\pm$ 0.12 & 86.94 $\pm$ 0.11 & 87.56 $\pm$ 0.10 & 88.23 $\pm$ 0.10 \\
		\rowcolor{LightCyan}
		+ \Last (LR) & 64.85 $\pm$ 0.20 & 81.81 $\pm$ 0.14 & 85.27 $\pm$ 0.12 & 87.19 $\pm$ 0.11 & 87.89 $\pm$ 0.10 & 88.45 $\pm$ 0.10 \\
		\midrule
		FEAT~\cite{Ye2018Learning}  & 66.78 $\pm$ 0.20 & 82.05 $\pm$ 0.14 & 85.15 $\pm$ 0.12 & 87.09 $\pm$ 0.11 & 87.82 $\pm$ 0.10 & 87.83 $\pm$ 0.10 \\
		\rowcolor{LightCyan}
		+ \Last (NC) & 67.33 $\pm$ 0.20 & 82.39 $\pm$ 0.14 & 85.64 $\pm$ 0.12 & 87.52 $\pm$ 0.11 & 88.26 $\pm$ 0.10 & 88.76 $\pm$ 0.10 \\
		\rowcolor{LightCyan}
		+ \Last (LR) & 67.35 $\pm$ 0.20 & 82.58 $\pm$ 0.14 & 85.99 $\pm$ 0.12 & 87.80 $\pm$ 0.11 & 88.63 $\pm$ 0.10 & 89.03 $\pm$ 0.10 \\
		\bottomrule
	\end{tabular}
	\label{tab:mini_cross_shot}
\end{table*}

\begin{table*}[t]
	\centering
	\caption{5-Way $K$-shot classification accuracy and 95\% confidence interval on {\it tiered}ImageNet, where $K\in\{1,5,10,20,30,50\}$. We apply \Last on ProtoNet and FEAT, and achieve stable improvements when the shot number becomes larger. For ``PT-EMB'', we apply the ProtoNet classification rule with the pre-trained feature extractor.}
	\begin{tabular}{c|cccccc}
		\addlinespace
		\toprule
		Setups $\rightarrow$ & 1     & 5     & 10    & 20    & 30    & 50 \\
		\midrule
		PT-EMB & 65.14 $\pm$ 0.23 & 84.65 $\pm$ 0.16 & 87.62 $\pm$ 0.14 & 89.24 $\pm$ 0.12 & 89.84 $\pm$ 0.12 & 90.26 $\pm$ 0.11 \\
		SimpleShot~\cite{Wang2019SimpleShot} & 70.51 $\pm$ 0.23 & 84.58 $\pm$ 0.16 & 87.64 $\pm$ 0.14 & 89.32 $\pm$ 0.12 & 89.77 $\pm$ 0.12 & 90.30 $\pm$ 0.11 \\
		\midrule
		ProtoNet~\cite{SnellSZ17Prototypical} & 68.23 $\pm$ 0.24 & 84.03 $\pm$ 0.16 & 86.28 $\pm$ 0.14 & 87.75 $\pm$ 0.13 & 88.33 $\pm$ 0.12 & 88.67 $\pm$ 0.11 \\
		\rowcolor{LightCyan}
		+ \Last (NC) & 69.02 $\pm$ 0.23 & 85.45 $\pm$ 0.16 & 88.32 $\pm$ 0.14 & 89.88 $\pm$ 0.12 & 89.97 $\pm$ 0.11 & 90.38 $\pm$ 0.11 \\
		\rowcolor{LightCyan}
		+ \Last (LR) & 68.95 $\pm$ 0.23 & 85.44 $\pm$ 0.16 & 88.04 $\pm$ 0.14 & 89.39 $\pm$ 0.12 & 89.94 $\pm$ 0.11 & 90.22 $\pm$ 0.11 \\
		\midrule
		FEAT~\cite{Ye2018Learning}  & 71.13 $\pm$ 0.23 & 84.79 $\pm$ 0.16 & 87.42 $\pm$ 0.14 & 89.13 $\pm$ 0.12 & 89.98 $\pm$ 0.11 & 90.11 $\pm$ 0.11 \\
		\rowcolor{LightCyan}
		+ \Last (NC) & 72.03 $\pm$ 0.23 & 85.66 $\pm$ 0.16 & 88.52 $\pm$ 0.14 & 90.06 $\pm$ 0.12 & 90.52 $\pm$ 0.11 & 90.94 $\pm$ 0.11 \\
		\rowcolor{LightCyan}
		+ \Last (LR) & 72.43 $\pm$ 0.23 & 85.82 $\pm$ 0.16 & 88.52 $\pm$ 0.14 & 90.06 $\pm$ 0.12 & 90.52 $\pm$ 0.11 & 91.06 $\pm$ 0.11 \\
		\bottomrule
	\end{tabular}
	\label{tab:tiered_cross_shot}
\end{table*}

\subsection{Larger-shot comparisons} 
Besides the benchmark results, we further investigate the scenarios when there are more shots in the support set for the novel classes (\ie, during meta-testing). For instance, $K=50$ is much larger than the usual few-shot learning scenario (where $K$=1 or $K$=5), but is still smaller than usual many-shot datasets like ImageNet~\cite{deng2009imagenet} for training a complex deep neural network.
We consider $K=\{1, 5, 10, 20, 30, 50\}$.
For meta-training on the base class data, we also consider $K=\{1, 5, 10, 20, 30, 50\}$. However, instead of setting the same $K$ for meta-training and meta-testing, we pick the best $K$ in meta-training for each $K$ of meta-testing, according to the accuracy on the meta-validation set. \emph{We note that, for a larger $K$ in meta-training, the query set size may be shrunk due to the memory constraint.} For example, for $K=50$, the number of query examples per class becomes $2$. For meta-testing, since no gradients need to be computed for the few-shot learner $\mathcal{A}$, we can still keep the query set size intact, \ie, $15$ examples per class.

\begin{figure}
	\centering
	\begin{minipage}[t]{0.49\linewidth}
		\centering
		\includegraphics[width=\textwidth]{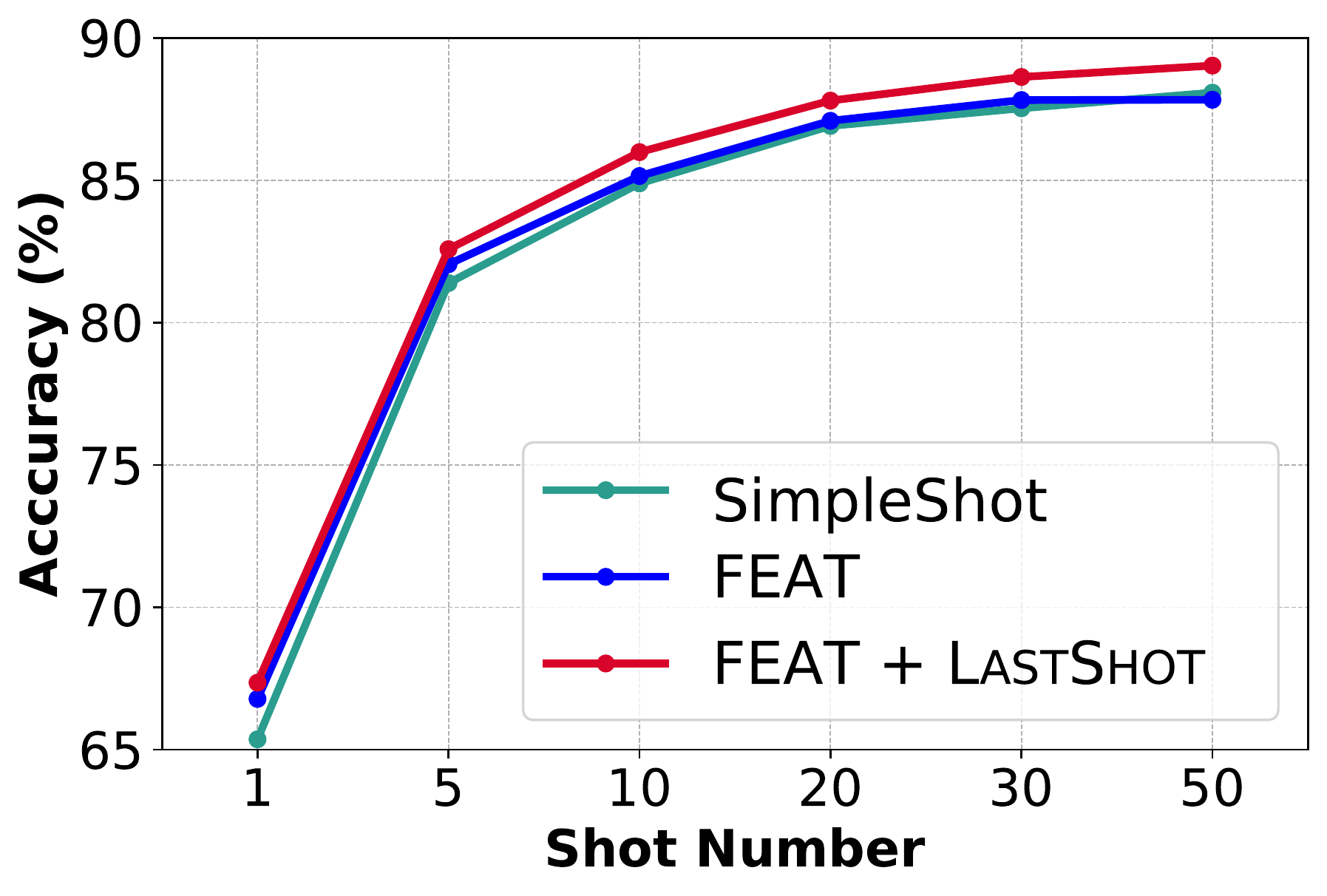}\\
	    \centering\mbox{(a) {\it mini}ImageNet}
	\end{minipage}
	\begin{minipage}[t]{0.49\linewidth}
		\centering
		\includegraphics[width=\textwidth]{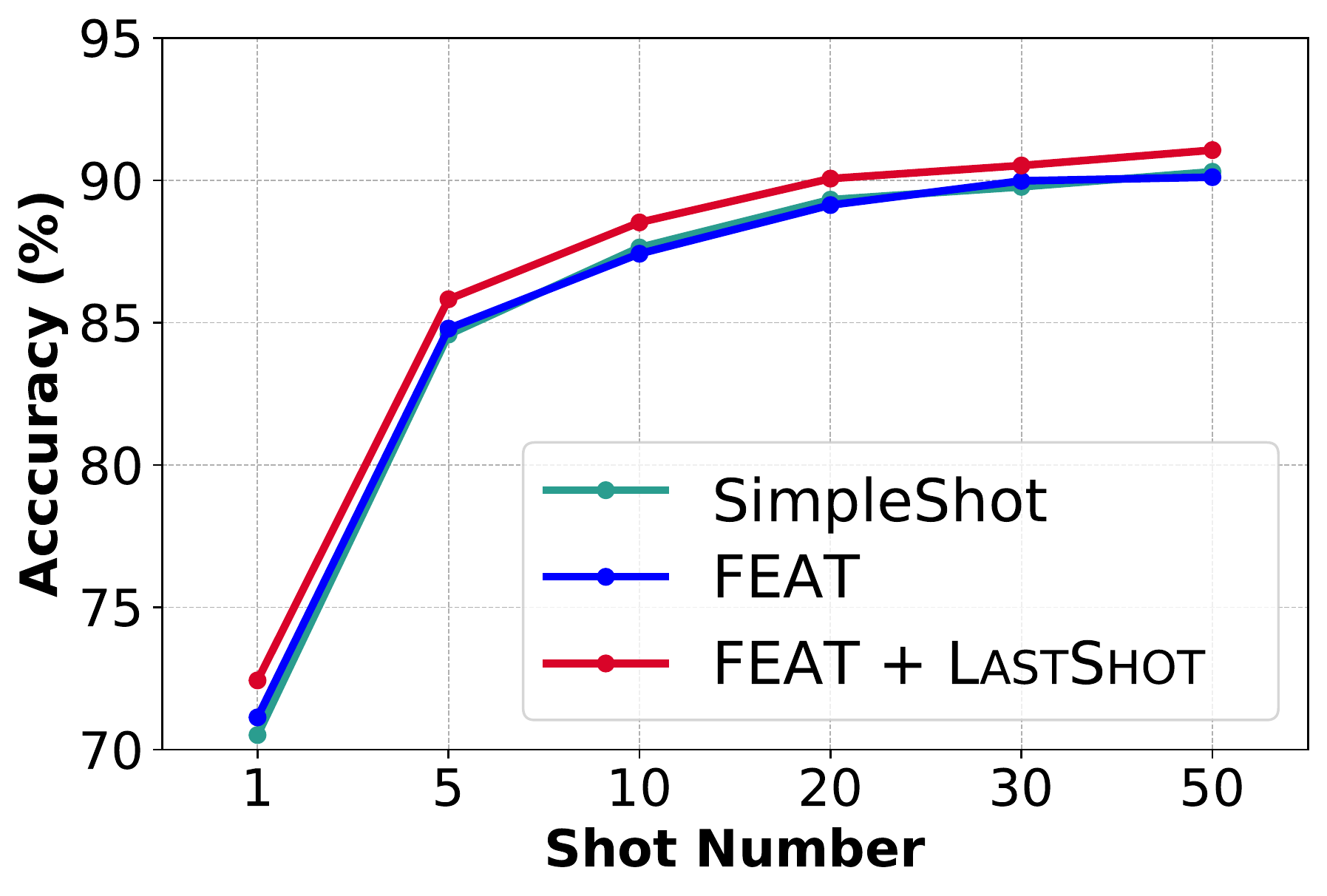}\\
	    \centering\mbox{(b) {\it tiered}ImageNet}
	\end{minipage}
	\caption{\textbf{The comparison of our \Last to existing methods.} We conduct 5-way (\ie, 5-class) classification experiments on {\it mini}ImageNet~\cite{VinyalsBLKW16Matching} and {\it tiered}ImageNet~\cite{ren2018learning} and report the accuracy under different numbers of shots.
	We consider two baselines: FEAT~\cite{Ye2018Learning} is a representative meta-learning based approach; SimpleShot~\cite{Wang2019SimpleShot} is a nearest centroid classifier whose features are learned by directly training a multi-class classifier using all the base class data.
	FEAT outperforms SimpleShot with smaller shot numbers but falls behinds with larger shots.
    Our proposed {\Last} improves FEAT for all settings.}
	\label{fig:shot_change_feat}
\end{figure}

The results are shown in \autoref{tab:mini_cross_shot} and \autoref{tab:tiered_cross_shot}: \Last can consistently improve the corresponding meta-learning method at all the different shot settings. From the further analysis in \autoref{fig:shot_change}, we see that the strength of conventional meta-learning decreases when $K$ increases. For instance, 
the improvement by the meta-learned method ProtoNet over the plainly-trained method ``PT-EMB'' diminishes when the number of shots becomes large. PT-EMB even outperforms ProtoNet when $K>5$. (Note that, both methods use exactly the same prediction rule.) In \autoref{fig:shot_change_feat}, we see a similar trend when comparing the best-performing meta-learning method FEAT and the best-performing non-meta-learning method SimpleShot. SimpleShot outperforms FEAT when $K = 50$. 

While one may argue that this is not surprising given that meta-learning is designed for small $K$, the fact that there seems to exist a \emph{turning point} that one must switch from meta-learning methods to non-meta-learning ones is notoriously annoying in practice, especially that such turning point is quite unstable (it is around 5 for ProtoNet but 50 for FEAT). It is desired to design learning mechanisms that can seamlessly combine the advantages of these two methods. 

We claim that \Last is one of such mechanisms. On the one hand, \Last is itself a meta-learning method, drawing few-shot tasks to train the few-shot learner. On the other hand, \Last explicitly uses the non-meta-learning based methods with a larger $K$ to construct the target classifier to teach the few-shot learner in the meta-training process.
The results on {\it mini}ImageNet and {\it tiered}ImageNet (in \autoref{tab:mini_cross_shot} and \autoref{tab:tiered_cross_shot}, respectively) further demonstrate this. Our {\Last} combined with FEAT consistently improves FEAT and outperforms SimpleShot at every $K$ values, making \Last the best-performing methods to be applied across a wide spectrum of $K$ values.
 
\subsection{Ablation studies}\label{sec:ablation}
\subsubsection{Advantage of \Last with a small query set}
We evaluate the influence of the query set size during meta-training. With a small query set, the few-shot learner only receives limited supervision. We investigate ProtoNet with \Last. As shown in \autoref{tab:shot_compare}, the classification accuracy of ProtoNet improves when more query instances are involved, especially from 1 to 5. With \Last, the few-shot learner performs much favorably. Even with 1 query instance per class, its performance can already outperform ProtoNet trained with 50 query instances but without \Last.  

\begin{table}[t]
	\centering
	\caption{The 1-Shot 5-Way classification accuracy and 95\% confidence interval with 10,000 trials on {\it mini}ImageNet. The tasks used for meta-training have different number of query instances $\{1, 15, 50\}$ per class.}
		\begin{tabular}{c|ccc}
			\addlinespace
			\toprule
			\# Query & 1 & 5 & 50 \\
			\midrule
			ProtoNet & 60.94 $\pm$  0.20 & 62.20 $\pm$  0.20 & 63.44 $\pm$  0.20 \\
			+ \Last(NC) & 63.78 $\pm$  0.20 & 64.16 $\pm$  0.20 & 64.50 $\pm$  0.20 \\
			+ \Last(LR) & 63.50 $\pm$  0.20 & 64.80 $\pm$  0.20 & 64.82 $\pm$  0.20 \\
			\bottomrule
	\end{tabular}
	\label{tab:shot_compare}
\end{table}

\begin{table}[t]
	\centering
	\caption{The 1/5-Shot 5-Way classification accuracy (plus 95\% confidence interval) with 10,000 trials on {\it mini}ImageNet. Based on FEAT, we compare the local teacher in our \Last (LR) with the teacher based on the pre-trained $B$-class classifier.}
		\begin{tabular}{c@{\;}|@{\;}cc@{\;}}
			\addlinespace
			\toprule
			& 1-Shot & 5-Shot \\
			\midrule
			FEAT  & 66.78  $\pm$  0.20 & 82.05  $\pm$  0.14 \\
			FEAT w/ $B$-class target & 66.89  $\pm$  0.20 & 82.10  $\pm$  0.14 \\
			FEAT + \Last(LR) & \bf 67.35  $\pm$  0.20 & {\bf 82.58  $\pm$  0.14} \\
			\bottomrule
	\end{tabular}
	\label{tab:compare_with_pt_teacher}
\end{table}

\begin{table}[t]
	\centering
	\caption{The 1/5-Shot 5-Way classification accuracy (plus 95\% confidence interval) with 10,000 trials on {\it mini}ImageNet and {\it tiered}ImageNet. Based on FEAT, we compare different implementations to query the target classifier.}
	\begin{tabular}{c@{\;}|@{\;}cc@{\;}}
			\addlinespace
			\toprule
			{\it mini}ImageNet & 1-Shot & 5-Shot \\
			\midrule
			FEAT  & 66.78  $\pm$  0.20 & 82.05  $\pm$  0.14 \\
			+ Autoaug & 65.22  $\pm$  0.20 & 80.13  $\pm$  0.14 \\
			+ Enlarge & 65.31  $\pm$  0.20 & 79.99  $\pm$  0.14 \\
			+ \Last(NC), Vanilla & 66.97  $\pm$  0.20 & {82.15  $\pm$  0.14} \\
			+ \Last(NC), Autoaug & 67.06  $\pm$  0.20 & {82.16  $\pm$  0.14} \\
			+ \Last(NC), Enlarge & \bf 67.33  $\pm$  0.20 & {\bf 82.39  $\pm$  0.14} \\
			\midrule\midrule
			{\it tiered}ImageNet & 1-Shot & 5-Shot \\
			\midrule
			FEAT  & 71.13  $\pm$  0.23 & 84.79  $\pm$  0.16 \\
			+ Autoaug & 68.11  $\pm$  0.23 & 82.48  $\pm$  0.16 \\
			+ Enlarge & 69.48  $\pm$  0.23 & 83.87  $\pm$  0.16 \\
			+ \Last(NC), Vanilla & 71.41  $\pm$  0.23 & {84.93  $\pm$  0.16} \\
			+ \Last(NC), Autoaug & 71.62  $\pm$  0.23 & {85.21  $\pm$  0.16} \\
			+ \Last(NC), Enlarge & \bf 72.03  $\pm$  0.23 & {\bf 85.66  $\pm$  0.16} \\
			\bottomrule
	\end{tabular}
	\label{tab:add_noise}
\end{table}

\subsubsection{Comparison on the target classifiers}
One intuitive way to construct the target classifier $h^\star$ is to train a $B$-class classifier and then select the corresponding $C$ logits according to the few-shot task at hand.
As shown in \autoref{tab:compare_with_pt_teacher}, using part of the $B$-class base classifier as the teacher could not introduce further improvement, which may result from the mismatch in class numbers and the poorly calibrated logit values.

\subsubsection{Strengthening the teacher or weakening the student?}
There are three ways to query $h^\star$, \ie, directly querying it, weakening the student (\eg, adding noise on the query set with autoaugment (Autoaug)~\cite{cubuk2019autoaugment}), or strengthening the teacher with enlarged inputs. We show the influence of the choices in \autoref{tab:add_noise}. The vanilla case means that we query $h^\star$ directly. By adding noises to the query set, we enlarge the learning signal between the teacher and the student; by enlarging the input images for the teacher, the teacher can provide higher-quality supervision. Both ways can improve the vanilla method. We empirically find that strengthening the teacher leads to the most stable improvement.

To verify if adding noise itself can improve meta-learning alone, we also apply Autoaug to FEAT. As shown in \autoref{tab:add_noise}, Autoaug, without \Last, can degrade the performance of FEAT significantly. Similarly, enlarging the input images without \Last cannot improve FEAT. 

\begin{table}[t]
	\centering
	\caption{The influence of $\lambda$ with {\Last}. We show the \{1,5,10,20,30,50\}-Shot 5-Way classification accuracy (plus 95\% confidence interval) with 10,000 trials on {\it mini}ImageNet.}
	\begin{tabular}{c|ccccccc}
		\addlinespace
		\toprule
		$\lambda$ & 0 & 0.001 & 0.01  & 0.1   & 1    & 10\\
    \midrule
    1-Shot & 67.08 & 66.82 & 67.13 & 66.74 & 66.91 & {\bf 67.33}\\
    5-Shot & 82.31 & 82.01 & {\bf 82.39} & 82.03 & 82.03 & 81.68 \\
    10-shot & 85.30 & 85.29 & {\bf 85.60} & 85.29 & 84.75 & 84.79 \\
    20-shot & 87.22 & \bf 87.44 & {\bf 87.44} & 87.16 & 86.55 & 86.51 \\
    30-shot & 87.82 & \bf 88.26 & {88.07} & 87.75 & 87.05 & 86.88 \\
    50-shot & 88.21 & \bf 88.76 & 88.58 & 88.24 & 87.69 & 87.34 \\
		\bottomrule
	\end{tabular}
	\label{tab:lambda_compare}
\end{table}

\subsubsection{Influence of the coefficient $\lambda$}
We investigate the influence of $\lambda$, the coefficient for the vanilla meta-learning objective in~\autoref{eq_Last},  when we apply \Last (NC) with FEAT. The larger the value of $\lambda$, the smaller the influence of the target classifier's supervision during meta-training. 
We show the \{1,5,10,20,30,50\}-Shot 5-Way classification accuracy on {\it mini}ImageNet in \autoref{tab:lambda_compare}.
We find that when there are more shots in a task, meta-learning with more supervision from the target classifier (with smaller $\lambda$) works better.

\subsubsection{Will a direct ensemble help?}
Since we construct a stronger teacher (not based on meta-learning) to supervise the few-shot learner, one may ask whether the stronger teacher itself can outperform the final few-shot learner trained with \Last. 
We note that, the strong teacher cannot be applied to the novel meta-testing tasks directly since we do not have ample examples for them to construct the strong teacher.
We therefore investigate an alternative question: can we achieve better few-shot accuracy by combining a non-meta-learning based and a meta-learning based FSL methods directly?
We investigate this idea by taking the ensemble of  their predictions.
In detail, for a meta-testing task, we apply both SimpleShot~\cite{Wang2019SimpleShot} and ProtoNet~\cite{SnellSZ17Prototypical} for classification, and we further average their predictions per example. We also apply the ensemble method to FEAT~\cite{Ye2018Learning}. The results are reported in \autoref{tab:ensemble}.
The ensemble approach gets marginal improvement compared to the vanilla meta-learning methods, while our \Last shows consistent improvements over the meta-learning methods with a single final model in various cases.

\begin{table}[t]
	\centering
	\caption{The 1/5-Shot 5-Way classification accuracy (plus 95\% confidence interval) with 10,000 trials on {\it mini}ImageNet. Based on ProtoNet and FEAT, we try to improve their performance by ensemble their predictions with a non-meta-learning based FSL method, \ie, SimpleShot~\cite{Wang2019SimpleShot}. The ensemble results are denoted by ``w/ Ensemble'' in the table.}
	\begin{tabular}{c@{\;}|@{\;}cc@{\;}}
			\addlinespace
			\toprule
			& 1-Shot & 5-Shot \\
			\midrule
			ProtoNet  & 62.39  $\pm$  0.20 & 79.74  $\pm$  0.14 \\
			w/ Ensemble  & 62.78  $\pm$  0.20 & 81.10  $\pm$  0.14 \\
			+ \Last(NC) & 64.16  $\pm$  0.20 & 81.23  $\pm$  0.14 \\
			\midrule
			FEAT  & 66.78  $\pm$  0.20 & 82.05  $\pm$  0.14 \\
			w/ Ensemble  & 66.85  $\pm$  0.20 & 81.76  $\pm$  0.14 \\
			+ \Last(NC) &  67.33  $\pm$  0.20 & 82.39  $\pm$  0.14 \\
			\bottomrule
	\end{tabular}
	\label{tab:ensemble}
\end{table}

\begin{table}[t]
  \centering
  \caption{The 1/5-Shot 5-Way classification accuracy (plus 95\% confidence interval) with 10,000 trials on {\it mini}ImageNet. Based on graph-based FSL methods, we try to improve their performance with \Last.}
    \begin{tabular}{c|cc}
    \addlinespace
    \toprule
          & 1-Shot & 5-Shot \\
    \midrule
    GCN   & 64.54  $\pm$  0.20 & 79.20  $\pm$  0.14 \\
    % \rowcolor{LightCyan}
    + \Last (NC)    & 65.17  $\pm$  0.20 & 81.36  $\pm$  0.14 \\
    % \rowcolor{LightCyan}
    + \Last (LR)   & 65.25  $\pm$  0.20 & 81.45  $\pm$  0.14 \\
    \bottomrule
    \end{tabular}
  \label{tab:gcn}
\end{table}

\subsubsection{Will \Last help graph-based FSL methods?}
We apply \Last to the GCN-based few-shot learning algorithm. We use \cite{Garcia2017Few} as the basic GCN-based algorithm (denoted as ``GCN''), following~\cite{Tseng2020Cross}. With ResNet-12 as the feature extractor, the 1/5-Shot 5-Way classification accuracy (plus 95\% confidence interval) with 10,000 trials on {\it mini}ImageNet is reported in~\autoref{tab:gcn}. Our \Last can further improve the GCN-based algorithm, arriving at  65.17\% w/ \Last (NC) and 65.25\% w/ \Last (LR), respectively.
The results indicate that \Last is general for meta-learning and can improve GCN-based algorithms.
 
\begin{table}[t]
  \centering
    \newcolumntype{g}{>{\columncolor{LightCyan}}c}
	\newcolumntype{f}{>{\columncolor{LightCyan}}l}
  \caption{The 1/5-Shot 5-Way classification accuracy (plus 95\% confidence interval) with 10,000 trials on {\it mini}ImageNet. Based on ProtoNet and FEAT, we try to improve their performance with \Last on ConvNet and WRN-28-10 backbones.}
  \tabcolsep 2pt
    \begin{tabular}{c|cc}
    \addlinespace
    \toprule
    ConvNet & 1-Shot & 5-Shot \\
          \midrule
    ProtoNet & 51.71  $\pm$  0.20 & 70.31  $\pm$  0.16 \\
    \rowcolor{LightCyan}
    +\Last(NC)    & 52.84  $\pm$  0.20 & 70.61  $\pm$  0.16 \\
    \rowcolor{LightCyan}
    +\Last(LR)    & 53.35  $\pm$  0.20 & 71.01  $\pm$  0.16 \\
    \midrule
    FEAT  & 54.62  $\pm$  0.20 & 71.95  $\pm$  0.16 \\
    \rowcolor{LightCyan}
    +\Last(NC)    & 56.19  $\pm$  0.20 & 72.62  $\pm$  0.16 \\
    \rowcolor{LightCyan}
    +\Last(LR)  & 56.19  $\pm$  0.20 & 72.35  $\pm$  0.16 \\
    \midrule\midrule
    WRN & 1-Shot & 5-Shot \\
    \midrule
    ProtoNet & 60.03  $\pm$  0.24 & 79.84  $\pm$  0.16 \\
    \rowcolor{LightCyan}
    +\Last(NC)  & 60.17  $\pm$  0.24 & 80.37  $\pm$  0.16 \\
    \rowcolor{LightCyan}
    +\Last(LR)   & 60.83  $\pm$  0.24 & 80.47  $\pm$  0.16 \\
    \midrule
    FEAT  & 65.09  $\pm$  0.24 & 81.74  $\pm$  0.16 \\
    \rowcolor{LightCyan}
    +\Last(NC)   & 65.89  $\pm$  0.24 & 82.35  $\pm$  0.16 \\
    \rowcolor{LightCyan}
    +\Last(LR)  & 65.86  $\pm$  0.24 & 82.27  $\pm$  0.16 \\
    \bottomrule
    \end{tabular}
  \label{tab:otherbackbone}
\end{table}

\subsubsection{Will \Last be compatible with other backbones?}

As shown in several recent works like~\cite{Dong2020NATS,Wang2019SimpleShot,Ye2018Learning}, the backbone architectures of the feature extractor can largely affect the resulting classification or FSL accuracy. In this subsection, 
we investigate whether the superiority of \Last extends to more backbones. We consider two widely used backbones, \ie, a four-layer convolution network and Wide ResNet-28-10~\cite{Zagoruyko2016WRN}.
The four-layer ConvNet contains four repeated blocks, and each block includes a convolutional layer, a Batch Normalization layer~\cite{IoffeS15BN}, a ReLU, and a Max pooling~\cite{VinyalsBLKW16Matching,SnellSZ17Prototypical}. We add a global average pooling layer at last to reduce the dimension of the embedding to 64. 
We follow~\cite{rusu2019meta,Ye2018Learning} and use the WRN-28-10 structure, which sets the depth to 28 and width to 10.
We study two meta-learning based algorithms, ProtoNet and FEAT. The 1/5-Shot 5-Way classification accuracy (plus 95\% confidence interval) with 10,000 trials on {\it mini}ImageNet is reported in~\autoref{tab:otherbackbone}. \Last improves ProtoNet and FEAT on both backbones.
\section{Experiments on Few-Shot Regression}\label{sec:regression}

Besides the experiments on few-shot classification in \autoref{sec:experiment}, we further follow \cite{FinnAL17Model} to study a one-dimensional regression problem to illustrate the effectiveness and applicability of our approach \Last.

\subsection{Datasets}
We generate synthetic, one-dimensional few-shot regression tasks with inputs $x$ sampled uniformly from $[-5,5]$. Every few-shot task is characterized by a true sine function parameterized by $(a,v,b)$:
\begin{align}
y = a\times\sin(v x+b) + \epsilon. \label{e_regreg}
\end{align}
In this sine curve, the amplitude $a$ is sampled uniformly from $[0, 2]$, the frequency $v$ is sampled uniformly from $[2,4]$, the bias $b$ is sampled uniformly from $[0, 2\pi]$, and $\epsilon$ is sampled from $0.3\times\mathcal{N}(0,1)$, where $\epsilon$ is an additional noise. 
Given $(a,v,b)$, we can then sample the support and query sets by first sampling $x\sim [-5, 5]$ and then obtaining the corresponding $y$ using \autoref{e_regreg}.
We note that, the sine curves defined in \autoref{e_regreg} have higher frequencies than those studied in~\cite{FinnAL17Model}. Thus, it requires more support set examples to learn a regression function.

\subsection{Evaluation and training protocols}
The goal is to build a regression function using the support set, such that it can predict the real-valued labels well for the corresponding query set.
We randomly sample $1,000$ meta-testing tasks (\ie, $1,000$ sin curves) for evaluation. From each, we randomly sample $K$ pairs of $(x, y)$ according to \autoref{e_regreg} to form the few-shot support set $\mathcal{D}'_\text{support}$, and another non-overlapping 100 pairs as the query set $\mathcal{D}'_\text{query}$. (In our experiments, we consider $K\in\{5, 50\}$.)
The performance is measured by the average Mean Square Error (MSE) and 95\% confidence intervals over the $1,000$ tasks.

We investigate training a meta-learning based few-shot learner $\mathcal{A}$ to output the regressor. 
Specifically, we model the regression function by a three-layer fully-connected neural network $f_{\vtheta}(x)$ as the feature extractor. The hidden and output dimensions are all set to $100$. We can learn such a feature extractor using nearly all the existing meta-learning methods. For example, with MAML~\cite{FinnAL17Model}, we will add a linear regressor $w$ on top. When embedding-based methods like ProtoNet~\cite{SnellSZ17Prototypical} are used, it is equivalent to learning a non-parametric regression
\begin{align}
\hat{y} = \sum_{(x', y')\in\mathcal{D}_\text{support}}\mathbf{Softmax}\left(-\|f_{\vtheta}(x) - f_{\vtheta}(x')\|_2^2\right)\times y',
\label{e_reg_ProtoNet}
\end{align}
where $\mathbf{Softmax}(\cdot)$ is taken over $K$ examples in $\mathcal{D}_\text{support}$.

We train $\mathcal{A}$ with at most $1,280,000$ few-shot tasks, each is associated with a specific $(a,v,b)$, a support set $\mathcal{D}_\text{support}$, and a query set $\mathcal{D}_\text{query}$. We pack every $32$ tasks into a mini-batch, and train $\mathcal{A}$ for $40,000$ iterations. We apply stochastic gradient descent (SGD) with a momentum of $0.9$ as the optimizer. We set the initial learning rate as $0.001$, and multiply it by $0.5$ every $160,000$ tasks. We perform early stopping using another $1,000$ validation tasks.

\begin{figure*}
	\begin{minipage}[h]{0.24\textwidth}
		\centering
		\includegraphics[width=\textwidth]{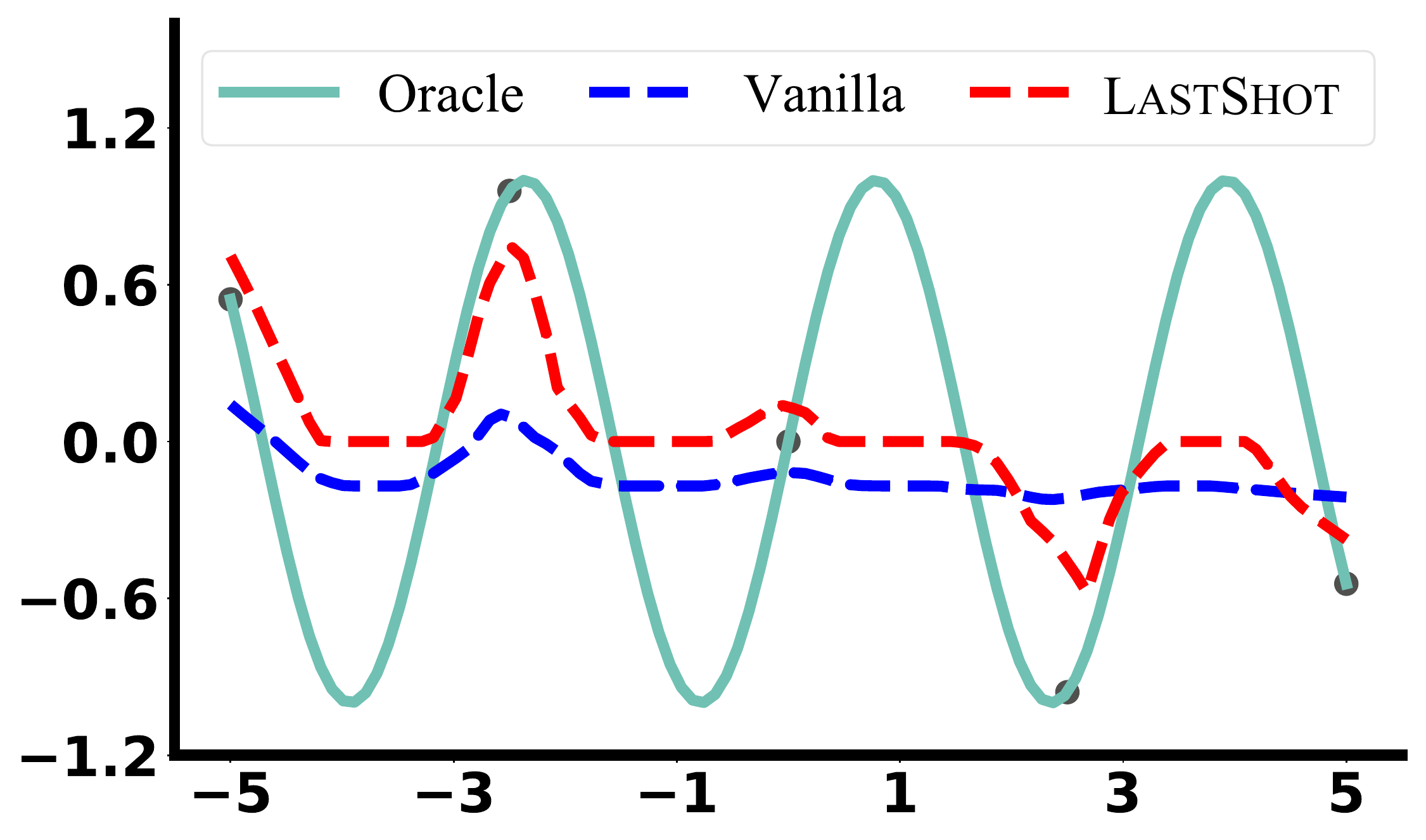}\\
		\centering\mbox{\small ({\it a}) {5-Shot, ProtoNet}}
	\end{minipage}
	\begin{minipage}[h]{0.24\textwidth}
		\centering
		\includegraphics[width=\textwidth]{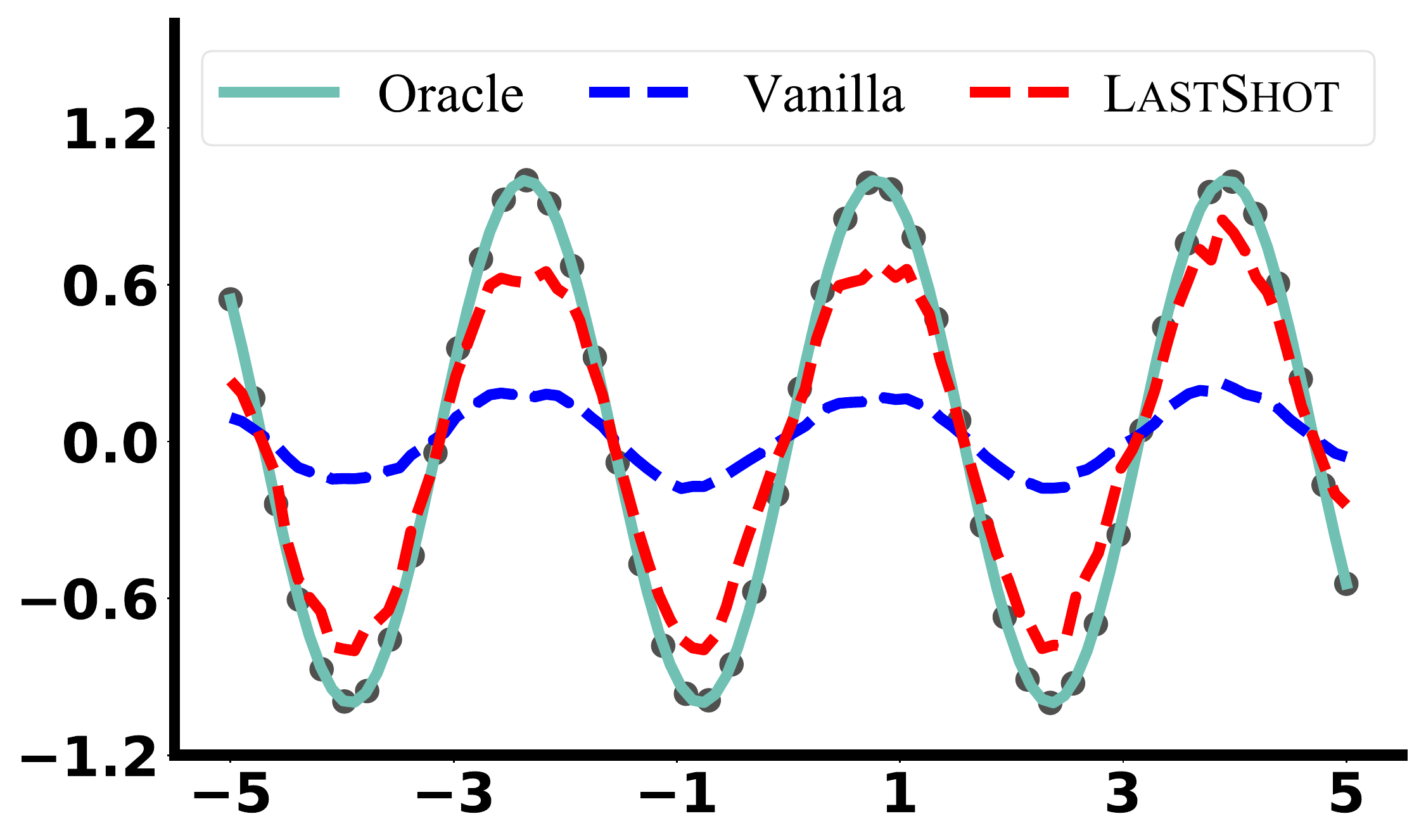}\\
		\centering\mbox{\small ({\it b}) {50-Shot, ProtoNet}}
	\end{minipage}
	\begin{minipage}[h]{0.24\textwidth}
		\centering
		\includegraphics[width=\textwidth]{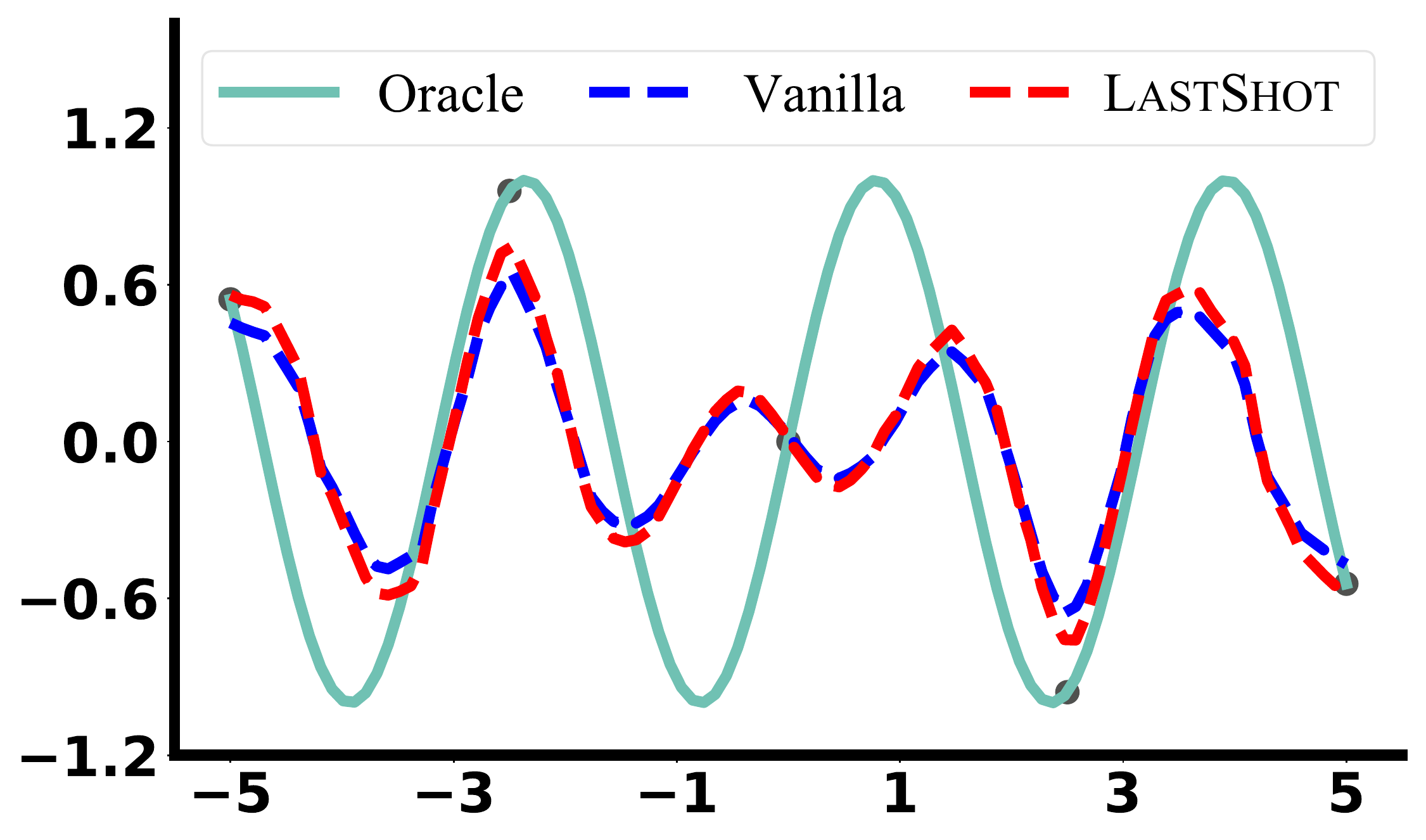}\\
		\centering\mbox{\small ({\it c}) {5-Shot, FEAT}}
	\end{minipage}
	\begin{minipage}[h]{0.24\textwidth}
		\centering
		\includegraphics[width=\textwidth]{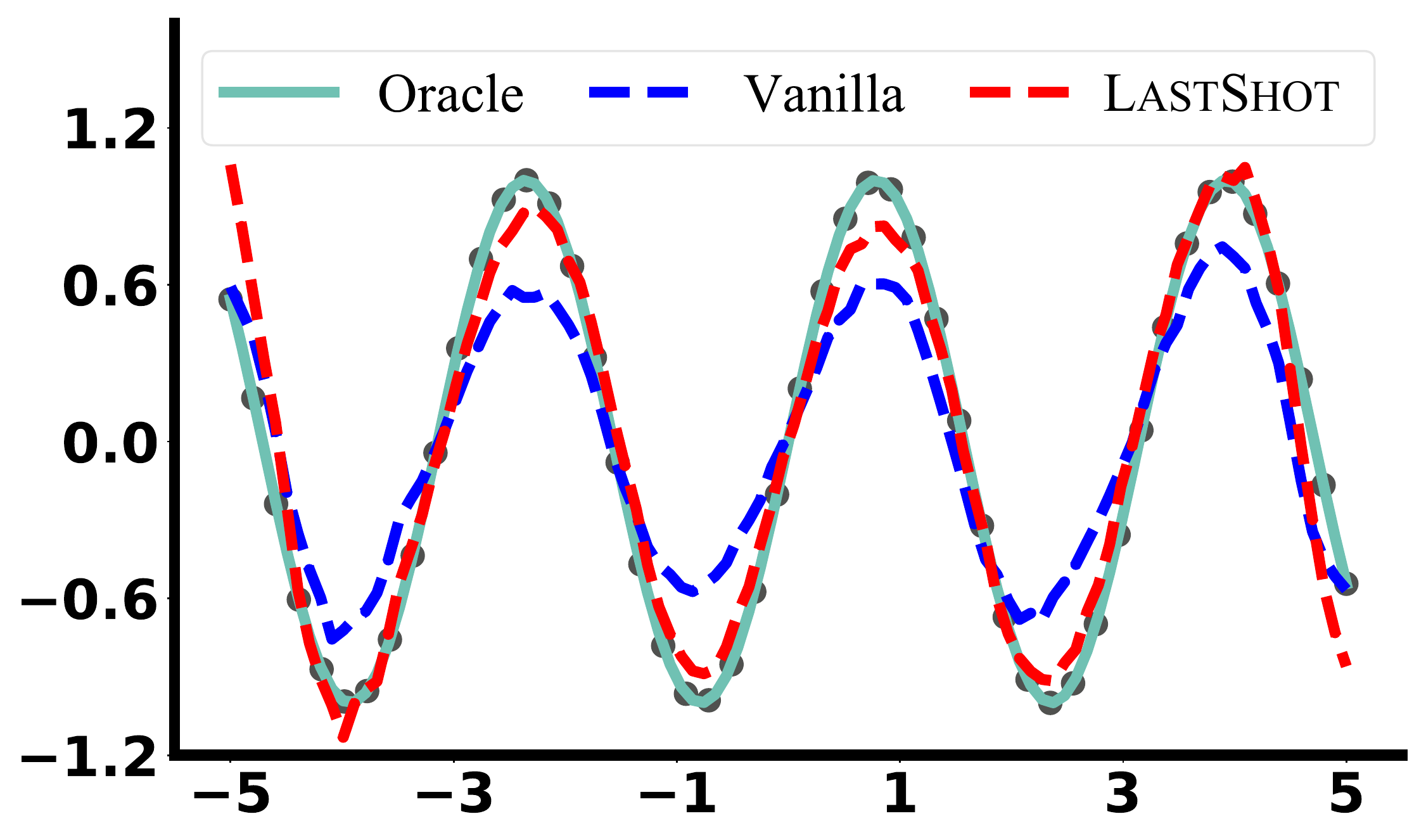}\\
		\centering\mbox{\small ({\it d}) {50-Shot, FEAT}}
	\end{minipage}

	\begin{minipage}[h]{0.24\textwidth}
		\centering
		\includegraphics[width=\textwidth]{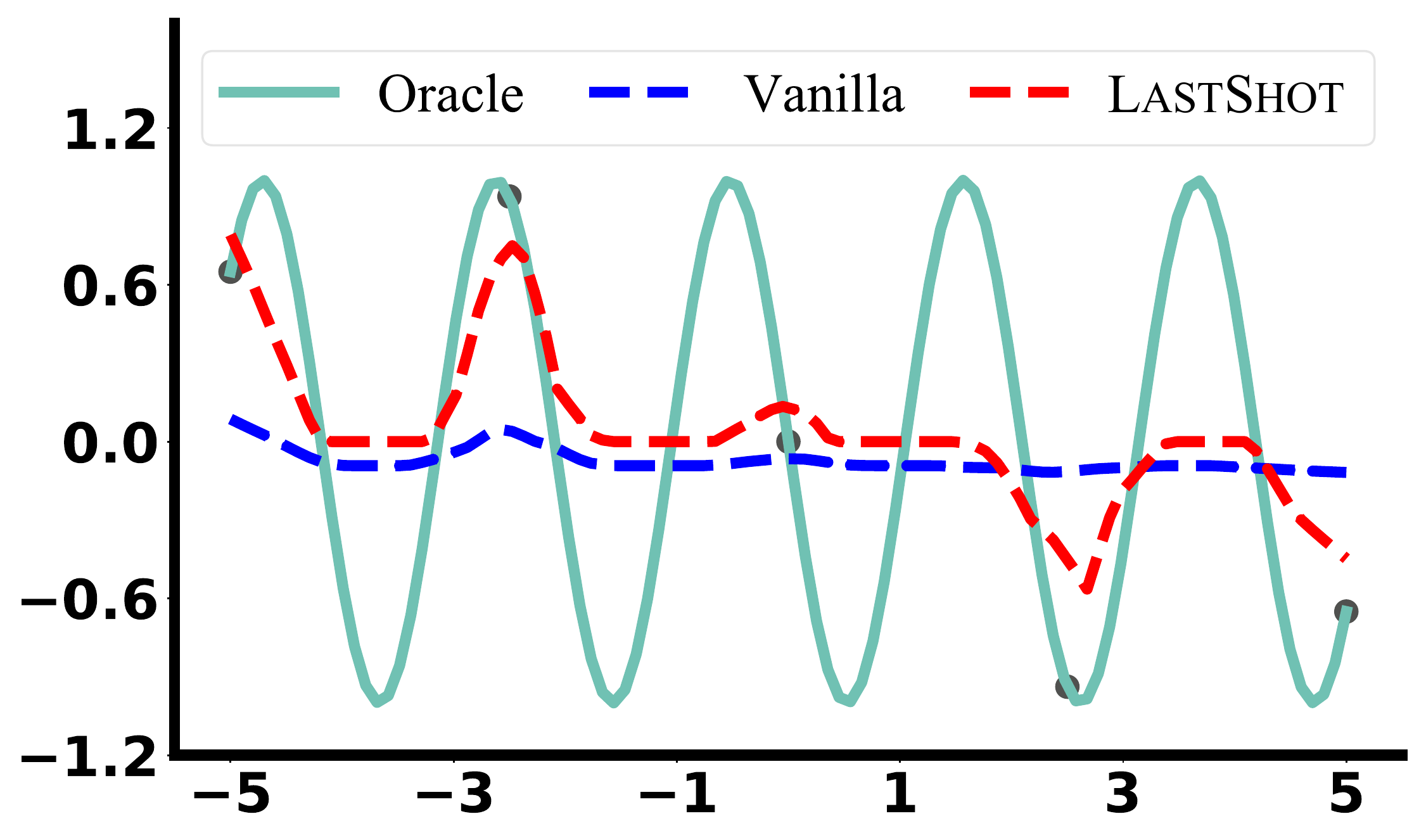}\\
		\centering\mbox{\small ({\it e}) {5-Shot, ProtoNet}}
	\end{minipage}
	\begin{minipage}[h]{0.24\textwidth}
		\centering
		\includegraphics[width=\textwidth]{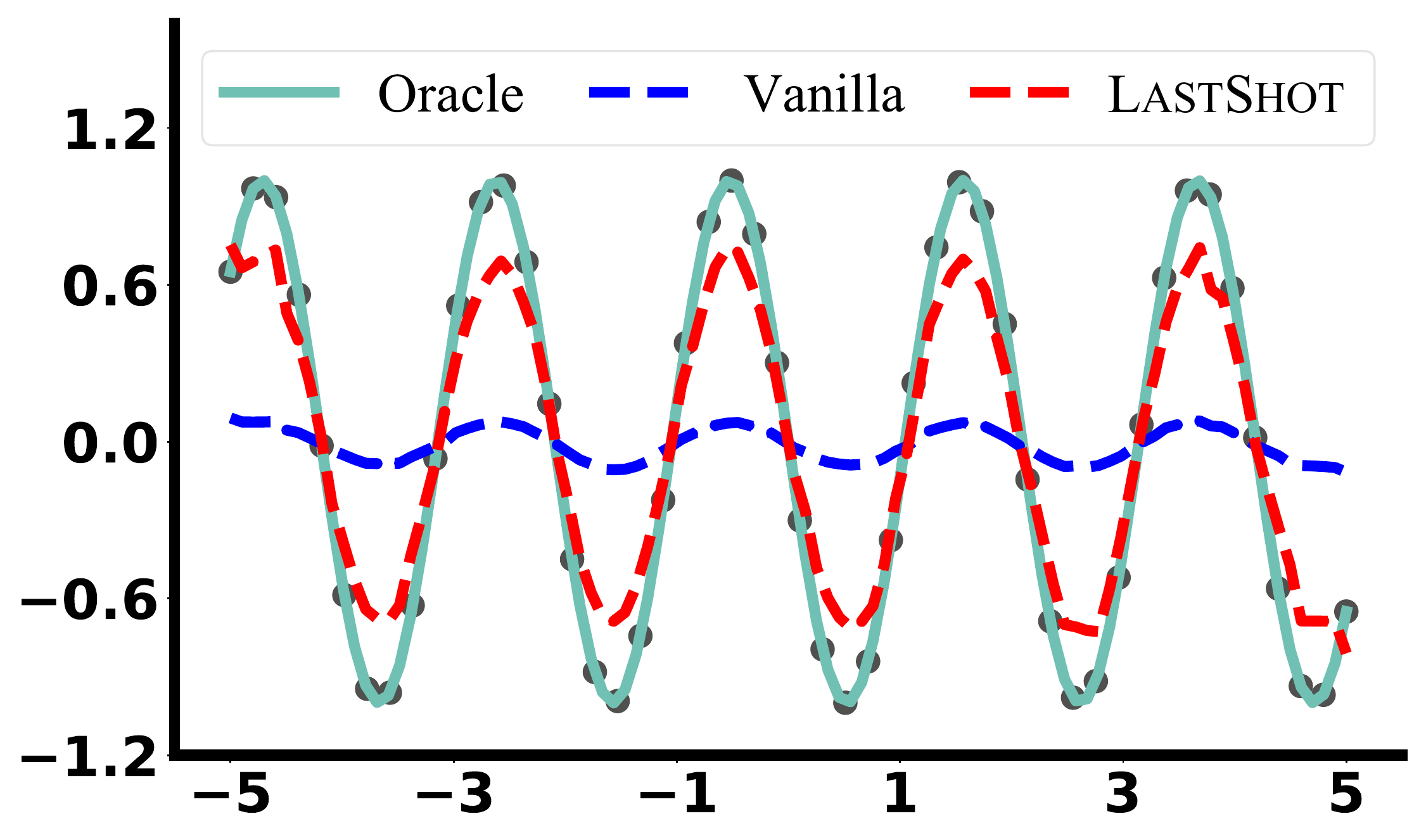}\\
		\centering\mbox{\small ({\it f}) {50-Shot, ProtoNet}}
	\end{minipage}
	\begin{minipage}[h]{0.24\textwidth}
		\centering
		\includegraphics[width=\textwidth]{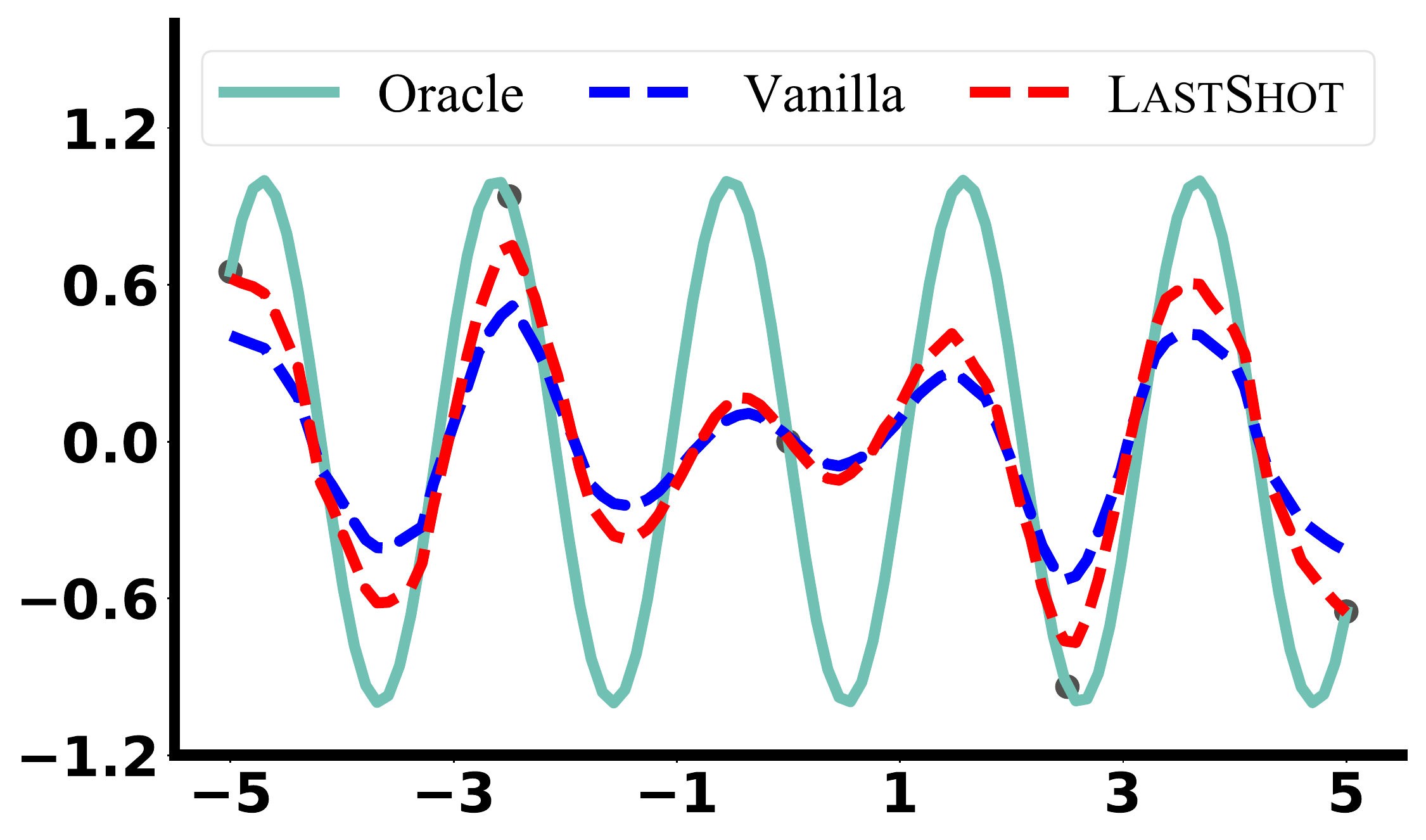}\\
		\centering\mbox{\small ({\it g}) {5-Shot, FEAT}}
	\end{minipage}
	\begin{minipage}[h]{0.24\textwidth}
		\centering
		\includegraphics[width=\textwidth]{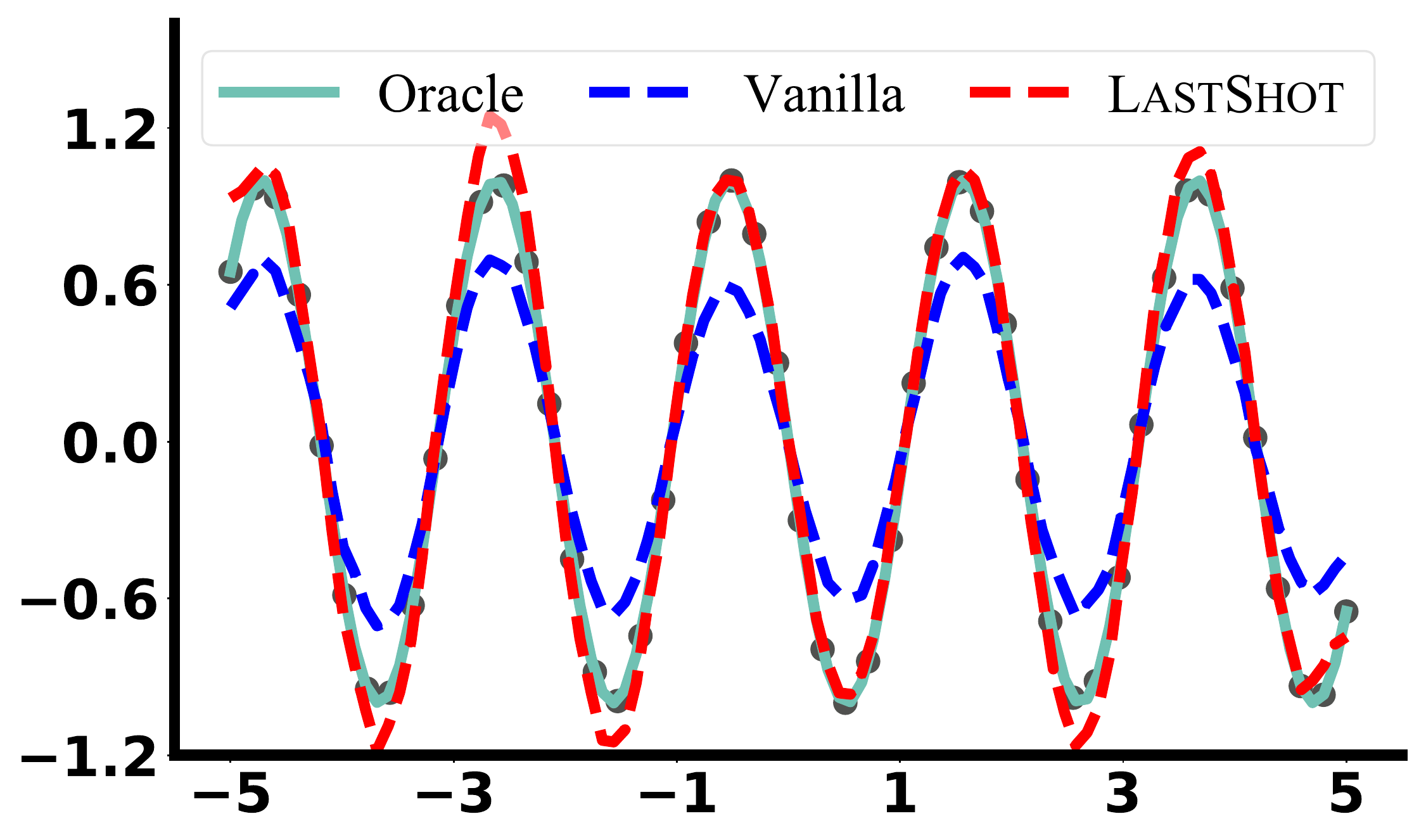}\\
		\centering\mbox{\small ({\it h}) {50-Shot, FEAT}}
	\end{minipage}

	\begin{minipage}[h]{0.24\textwidth}
		\centering
		\includegraphics[width=\textwidth]{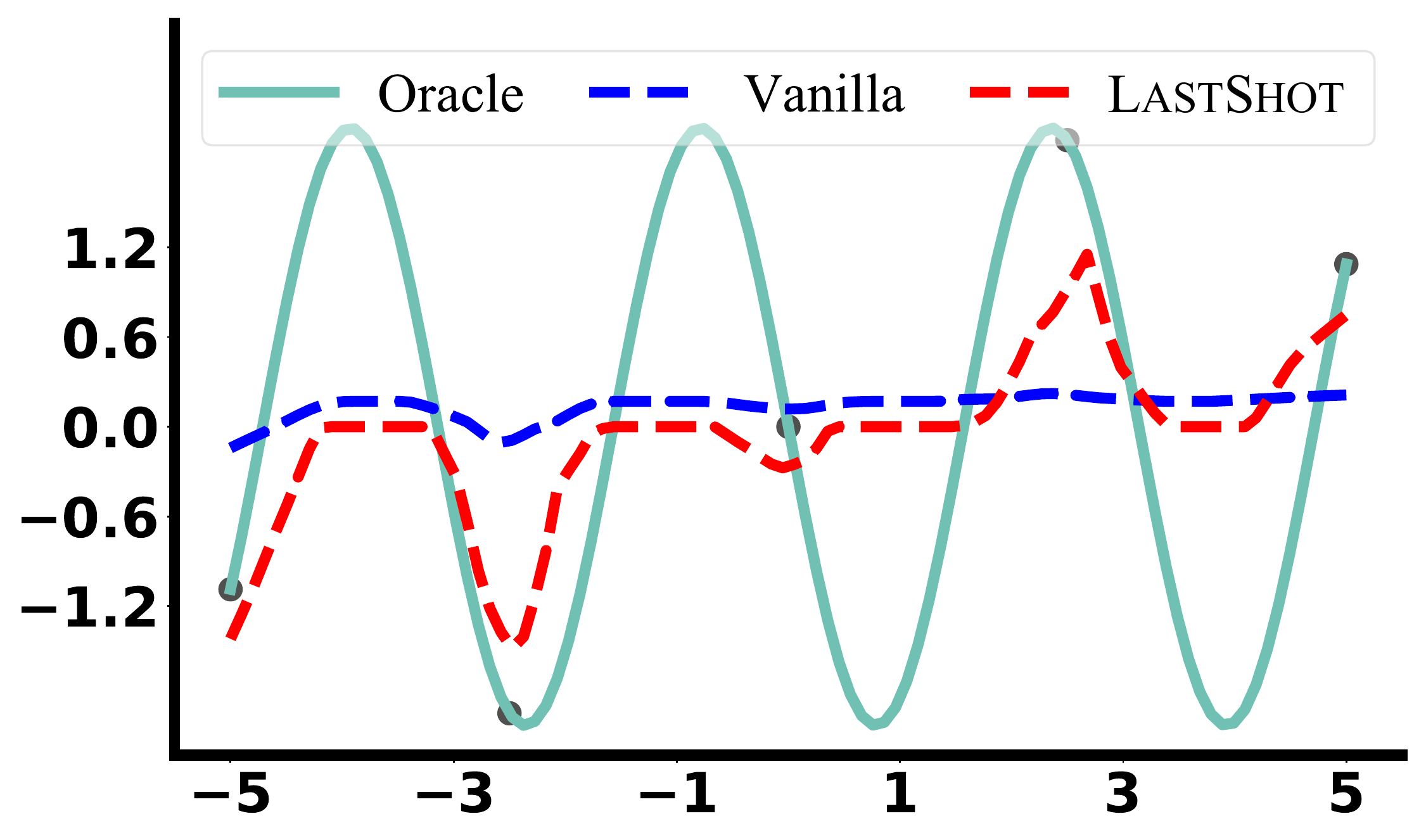}\\
		\centering\mbox{\small ({\it i}) {5-Shot, ProtoNet}}
	\end{minipage}
	\begin{minipage}[h]{0.24\textwidth}
		\centering
		\includegraphics[width=\textwidth]{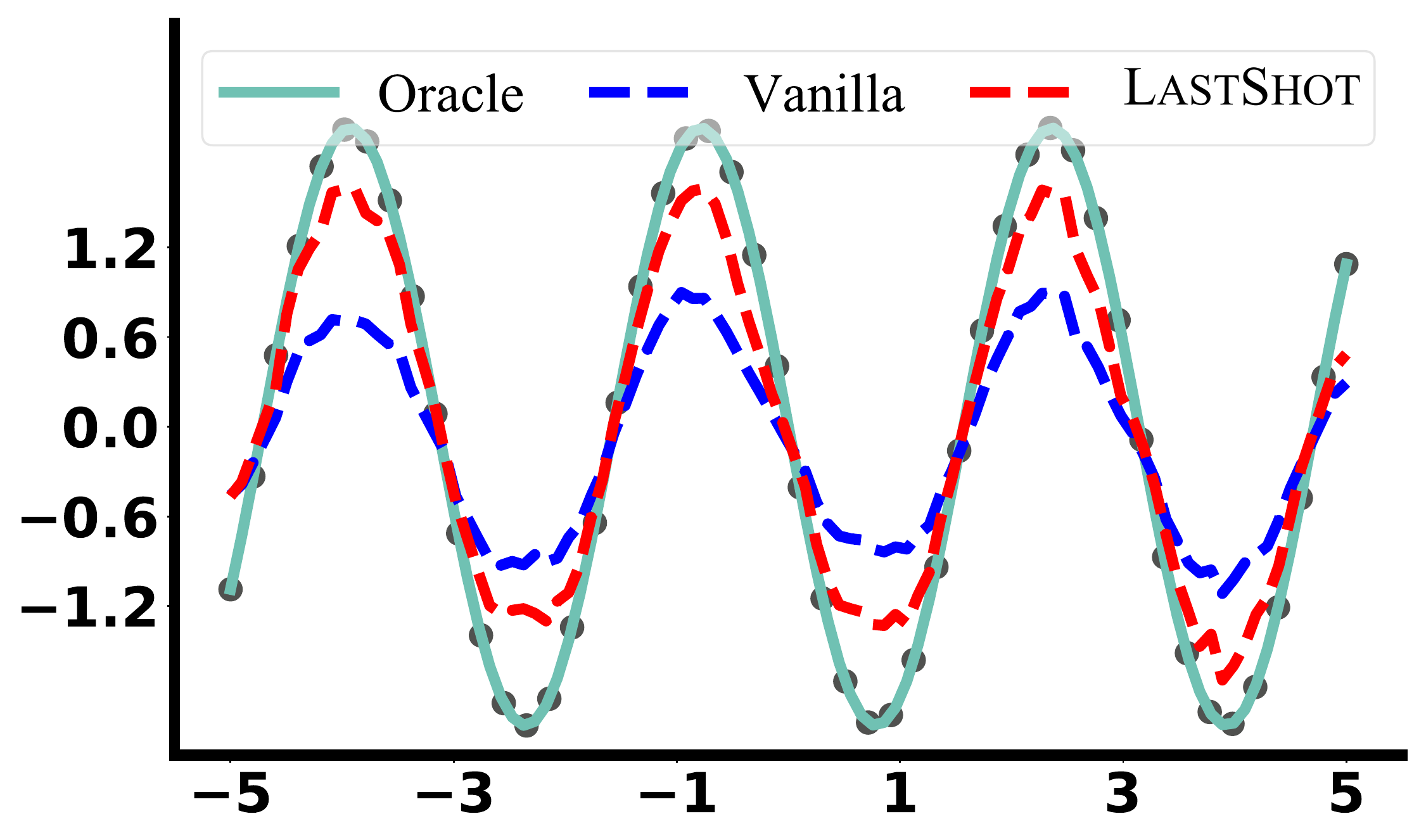}\\
		\centering\mbox{\small ({\it j}) {50-Shot, ProtoNet}}
	\end{minipage}
	\begin{minipage}[h]{0.24\textwidth}
		\centering
		\includegraphics[width=\textwidth]{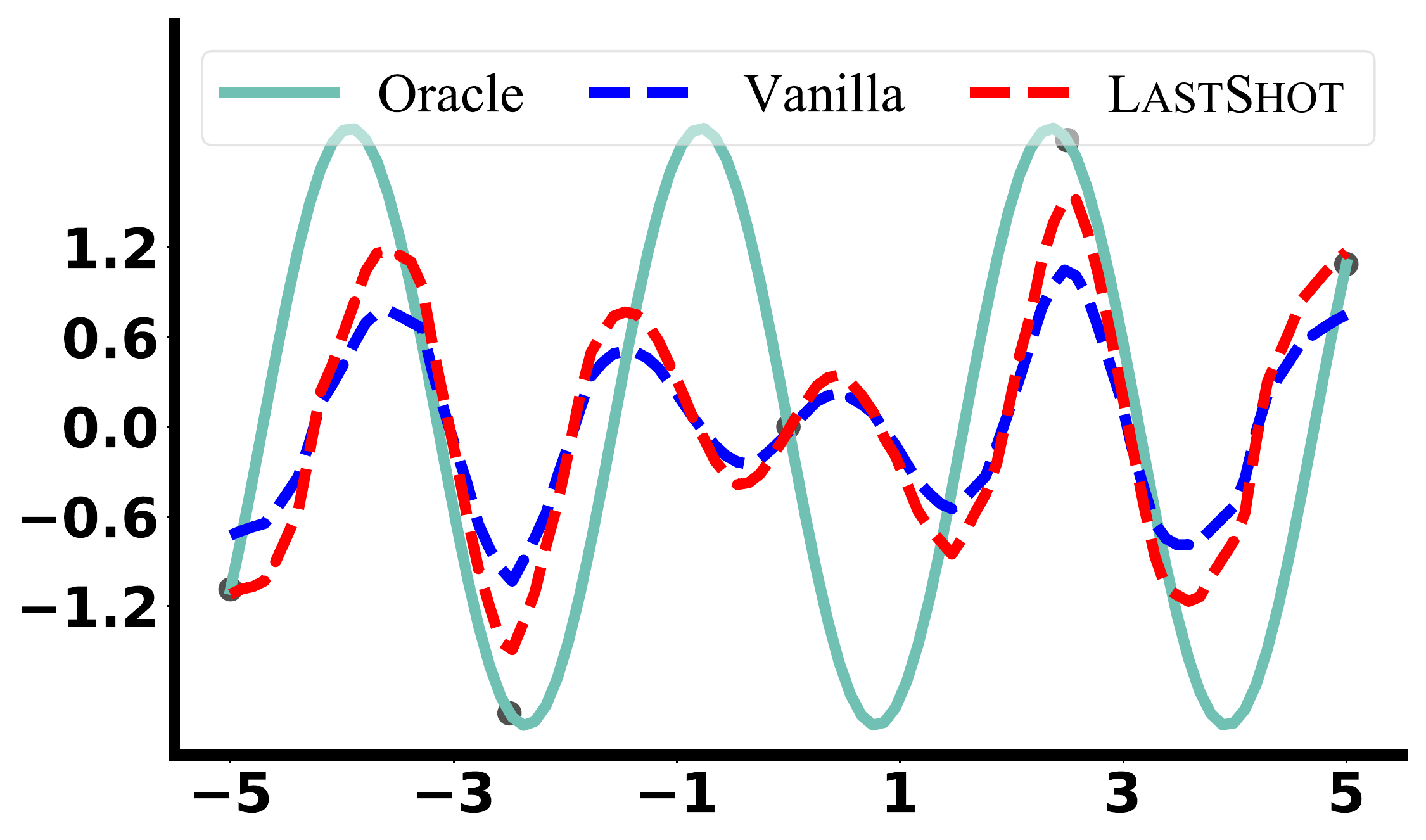}\\
		\centering\mbox{\small ({\it k}) {5-Shot, FEAT}}
	\end{minipage}
	\begin{minipage}[h]{0.24\textwidth}
		\centering
		\includegraphics[width=\textwidth]{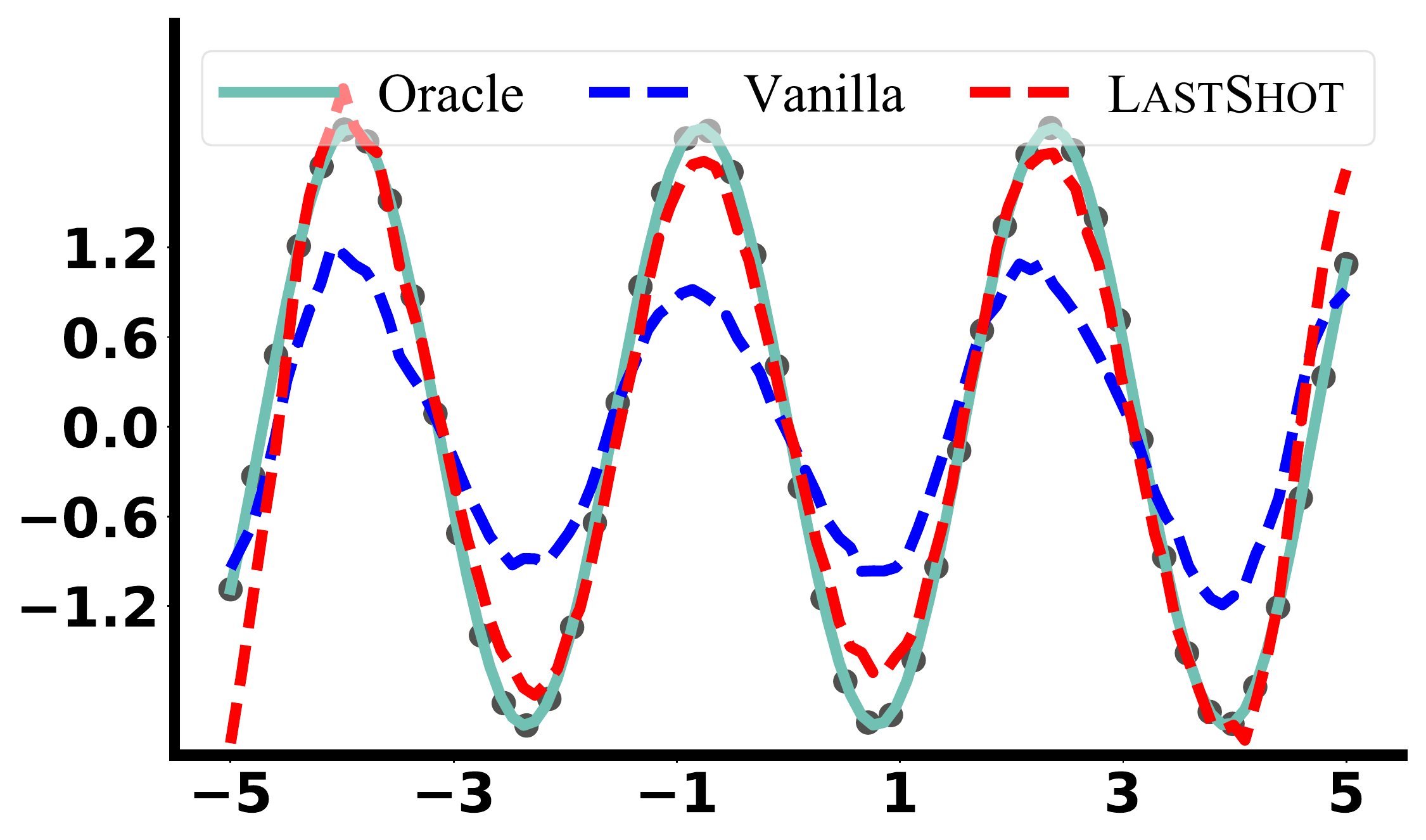}\\
		\centering\mbox{\small ({\it l}) {50-Shot, FEAT}}
	\end{minipage}
	\vskip-5pt
	\caption{Visualizing the 5-shot and 50-shot regression tasks on three curves (each takes a row). The ground-truth curves (from top to bottom) are $y=\sin(2x)$, $y=\sin(3x+\pi)$, and $y=2\sin(2x+\pi)$, respectively.
	Black points are the training examples in the support set. ``Oracle'' ({cyan}) denotes the ground-truth curve in the range $[-5,5]$ (we draw training examples without additional noise $\epsilon$ for visualization clarity).
	``Vanilla'' ({blue}) means the predicted curve by ProtoNet or FEAT without \Last. 
	Our {\Last} ({red}) is implemented on top of each ``Vanilla'' method and can produce much closer curves to the oracle ones.}\label{fig:regression_visualization}
	\vskip-5pt
\end{figure*}

\subsection{{\Last} for regression}
To apply {\Last} for few-shot regression, we must (a) construct the target regressor $h^\star$ for each few-shot regression task, and (b) define the suitable $\ell_\Last$ for supervision.

For the former, to efficiently construct $h^\star$ without sampling extra data from each few-shot task, we discretize the ranges of $a\in[0, 2]$, $v\in[2,4]$, and $b\in[0, 2\pi]$ each by $0.1$ to construct a set of ``anchor'' tasks. For each of them, we can sample many-shot examples (\eg, $1,000$ in our experiments) to construct a many-shot regressor. Here we train a neural network regressor using the regularized least square loss (\ie, the loss used in ridge regression) for each anchor task, whose hyper-parameter is tuned by cross-validation.
Then during meta-training, given a few-shot task and its associated $(a, v, b)$, we retrieve the anchor task at the same grid to be the target model $h^\star$ --- the rationale here is that sine curves with similar $(a, v, b)$ are shaped similarly.

With the target regressor $h^\star$, we realize $\ell_\Last$ (cf. \autoref{eq_Last}) by a weighted square loss~\cite{Saputra2019Distilling},
\begin{align}
\sum_{(x,y) \in\mathcal{D}_\text{query}} & \ell_\Last(\mathcal{A}(\mathcal{D}_\text{support})(x), h^\star(x)) = \nonumber \\
\sum_{(x,y)\in\mathcal{D}_\text{query}} & \mathbf{Softmax}\left(-(h^\star(x) - y)^2\right) \\ 
& \times\left(h^\star(x) -  \mathcal{A}(\mathcal{D}_\text{support})(x)\right)^2\;, \nonumber
\end{align}
in which $\mathbf{Softmax}$ is taken over the examples in $\mathcal{D}_\text{query}$. The $\mathbf{Softmax}$ measures how well the target regressor $h^\star$ performs on a query example --- we measure this because the retrieved anchor regressor does not exactly match the few-shot task at hand. The better $h^\star$ performs (\ie, a smaller error between $h^\star(x)$ and $y$), the larger the score is after $\mathbf{Softmax}$. We then use these $\mathbf{Softmax}$ scores to perform weight average over the error between the few-shot regressor $A(\mathcal{D}_\text{support})(x)$ and the target regressor $h^\star(x)$ for each query example $x$. Namely, the few-shot learner pays more attention to those query examples that $h^\star(x)$ truly performs well on.

\subsection{\bf Results}
We compare {\Last} with ProtoNet~\cite{SnellSZ17Prototypical}, MAML~\cite{FinnAL17Model}, MetaOptNet~\cite{Lee2019Meta}, and FEAT~\cite{Ye2018Learning} in \autoref{tab:regression}.
Specifically for MetaOptNet, we implement its top-layer classifier as ridge regression and tune the hyper-parameter.
We make the following observations. First, predicting a sine curve is difficult given few-shot examples\footnote{One-dimensional sine functions have an infinite VC dimension.}, and all methods perform better when there are more shots. Second, FEAT and MetaOptNet, being more advanced meta-learning methods, outperform MAML and ProtoNet. 
Third and most importantly, {\Last} can notably improve each of the baseline methods.

We further visualize the few-shot learners' performance on three few-shot tasks whose true, oracle curve are $y=\sin(2x)$, $y=\sin(3x+\pi)$, and $y=2\sin(2x+\pi)$ in \autoref{fig:regression_visualization}. Specifically, we consider both $5$ and $50$ shots cases, and plot the oracle curve ({cyan}), predicted curve without \Last ({blue}), and predicted curve with \Last ({red}). We see that the predicted curves with \Last can better match the oracle ones. 
It is worth noting that for each of these three curves, there is more than one period in the range. Therefore, it is difficult to fit the curve when there are merely five examples: there can be multiple feasible solutions. We surmise that this is why ProtoNet creates a nearly horizontal curve. By learning with a stronger teacher via {\Last}, ProtoNet can generate more faithful curves in both 5-shot and 50-shot scenarios. 

\begin{table}[tbp]	
	\centering
	 \newcolumntype{g}{>{\columncolor{LightCyan}}c}
	\newcolumntype{f}{>{\columncolor{LightCyan}}l}
	\caption{Mean Square Error (MSE) and 95\% confidence interval over 1000 $\{5, 50\}$-shot regression tasks.  The lower the better.
}
\vskip-5pt
	\tabcolsep 4pt
	\begin{tabular}{l|cc}
		\addlinespace
		\toprule
		MSE   & 5-Shot & 50-Shot \\
		\midrule
		MAML{~\cite{FinnAL17Model}}  &  0.693$\pm$0.034  &  0.303$\pm$0.016\\
		\midrule
		ProtoNet{~\cite{SnellSZ17Prototypical}} & 0.681$\pm$0.031  &  0.299$\pm$0.013\\
		\rowcolor{LightCyan}
		+ \Last &  0.630$\pm$0.030  &  0.294$\pm$0.011\\
		\midrule
		MetaOptNet{~\cite{Lee2019Meta}} & 0.622$\pm$0.029  &  0.162$\pm$0.004\\
		\rowcolor{LightCyan}
		+ \Last &  0.616$\pm$0.022  & \bf  0.159$\pm$0.004\\
		\midrule
		FEAT{~\cite{Ye2018Learning}} & 0.491$\pm$0.026  &  0.189$\pm$0.006\\
		\rowcolor{LightCyan}
		+ \Last &  \bf 0.479$\pm$0.023  &  0.167$\pm$0.004\\
		\bottomrule
	\end{tabular}
	\label{tab:regression}
	\vskip-5pt
\end{table}

\section{Conclusion}
\label{s_disc}
We present a novel meta-learning paradigm to introduce rich supervision for few-shot learning. By coupling a few-shot task in the meta-training phase with a target classifier that is trained with ample examples, the few-shot learner can receive additional training signals from the target classifier besides the query set instances. 
We propose ways to efficiently construct and leverage the target classifier, leading to a plug-and-play method \Last, which is compatible with various meta-learning based few-shot learning methods.
On multiple datasets, \Last achieves consistent improvement against baseline meta-learning methods. Importantly, \Last also maintains robust performance across a wider range of shot numbers --- making meta-learning not only work well on the few-shot but also many-shot settings.

\ifCLASSOPTIONcompsoc
\section*{Acknowledgments}
\else
\section*{Acknowledgment}
\fi
This research is supported by National Key R\&D Program of China (2020AAA0109401), Nanjing University-Huawei Joint Research Program, NSFC (62006112, 61921006, 62176117), Collaborative Innovation Center of Novel Software Technology and Industrialization, NSF of Jiangsu Province (BK20200313), NSF IIS-2107077, NSF OAC-2118240, NSF OAC-2112606, and the OSU GI Development funds.
We are thankful for the generous support of computational resources by Ohio Supercomputer Center and AWS Cloud Credits for Research.

\ifCLASSOPTIONcaptionsoff
\newpage
\fi

{\small
\bibliographystyle{IEEEtran}
\bibliography{main}
}

\ifCLASSOPTIONcaptionsoff
\newpage
\fi

\end{document}